\documentclass[sn-mathphys-num]{sn-jnl}

\usepackage{makecell}
\usepackage[table]{xcolor} 
\usepackage{graphicx}%
\usepackage{multirow}%
\usepackage{amsmath,amssymb,amsfonts}%
\usepackage{amsthm}%
\usepackage{mathrsfs}%
\usepackage[title]{appendix}%
\usepackage{xcolor}%
\usepackage{textcomp}%
\usepackage{wrapfig}
\usepackage{manyfoot}%
\usepackage{booktabs}%
\usepackage{algorithm}%
\usepackage{algorithmicx}%
\usepackage{algpseudocode}%
\newcommand{\myparatight}[1]{\smallskip\noindent{\bf {#1}:}~}
\usepackage{listings}%
\usepackage{subcaption}
\usepackage{array} 
\usepackage{booktabs} 
\usepackage{lipsum}
\usepackage{enumerate}

\theoremstyle{thmstyleone}%
\theoremstyle{thmstyletwo}%

\theoremstyle{thmstylethree}%
\definecolor{greyL}{RGB}{230,248,255}
\definecolor{mypink}{RGB}{255, 204, 211}
\definecolor{mybluegray}{RGB}{204, 209, 224}

\usepackage{wrapfig} 
\definecolor{cvprblue}{rgb}{0.21,0.49,0.74}

\begin{document}

\title[Article Title]{DSADF: Thinking Fast and Slow for Decision Making}


\author[1]{\fnm{Zhihao} \sur{Dou}}\email{zdou713@gmail.com}
\equalcont{These authors contributed equally to this work.}

\author[2]{\fnm{Dongfei} \sur{Cui}}\email{cuidongfei31@gmail.com}
\equalcont{These authors contributed equally to this work.}

\author*[3]{\fnm{Jun} \sur{Yan}}\email{yanjun@ieee.org}
\equalcont{These authors contributed equally to this work.}

\author[4]{\fnm{Weida} \sur{Wang}}\email{2151300@tongji.edu.cn}

\author[5]{\fnm{Benteng} \sur{Chen}}\email{2021090076@buct.edu.cn}

\author[6]{Haoming Wang} \email{wfrank0222@gmail.com}

\author[7]{Zeke Xie} \email{zekexie@hkust-gz.edu.cn}

\author*[8]{\fnm{Shufei} \sur{Zhang}}\email{zhangshufei@pjlab.org.cn}

\affil[1]{\orgdiv{Pratt School of Engineering}, \orgname{Duke University}, \orgaddress{\street{103 Allen Bldg}, \city{Durham}, \postcode{27708}, \state{North Carolina}, \country{USA}}}

\affil[2]{\orgdiv{School of Computer Science}, \orgname{Northeast Electric Power University}, \orgaddress{\street{No. 169, Changchun Road, Chuanying District,}, \city{Jilin}, \postcode{132012}, \state{Jilin}, \country{China}}}

\affil[3]{\orgdiv{Department of Information and Communication Engineering}, \orgname{Tongji University}, \orgaddress{\street{No. 4800, Caoan Highway}, \city{Shanghai}, \postcode{201804}, \state{Shanghai}, \country{China}}}

\affil[4]{\orgdiv{School of Computer Science and Technology}, \orgname{Tongji University}, \orgaddress{\street{No. 4800, Caoan Highway}, \city{Shanghai}, \postcode{201804}, \state{Shanghai}, \country{China}}}

\affil[5]{\orgdiv{School of International Education}, \orgname{Beijing University of Chemical Technology}, \orgaddress{\street{No. 15 Beisanhuan East Road}, \city{Chaoyang District}, \postcode{100029}, 
\state{Beijing}, \country{China}}}

\affil[6]{\orgdiv{Department of Education Information Technology, Faculty of Education}, \orgname{East China Normal University}, \orgaddress{\street{3667 North Zhongshan Road, Changfeng Xincun Subdistrict, Putuo District}, \city{Pudong New Area}, \postcode{200062}, \state{Shanghai}, \country{China}}}

\affil[7]{\orgdiv{Information Hub}, \orgname{Hong Kong University of Science and Technology (Guangzhou)}, \orgaddress{\street{Duxue Road No.1, Nansha}, \city{Guangzhou}, \postcode{511453}, 
\state{Guangdong}, \country{China}}}

\affil[8]{\orgdiv{Shanghai Artificial Intelligence Laboratory }, \orgname{Chinese Academy of Science}, \orgaddress{\street{No. 1 Xueyang Road}, \city{Pudong New Area}, \postcode{201304}, \state{Shanghai}, \country{China}}}


\abstract{
Although Reinforcement Learning (RL) agents are effective in well-defined environments, they often struggle to generalize their learned policies to dynamic settings due to their reliance on trial-and-error interactions. Recent work has explored applying Large Language Models (LLMs) or Vision Language Models (VLMs) to boost the generalization of RL agents through policy optimization guidance or prior knowledge. However, these approaches often lack seamless coordination between the RL agent and the foundation model, leading to unreasonable decision-making in unfamiliar environments and efficiency bottlenecks. Making full use of the inferential capabilities of foundation models and the rapid response capabilities of RL agents and enhancing the interaction between the two to form a dual system is still a lingering scientific question. To address this problem, we draw inspiration from Kahneman's theory of fast thinking (System 1) and slow thinking (System 2), demonstrating that balancing intuition and deep reasoning can achieve nimble decision-making in a complex world. In this study, we propose a \textbf{Dual-System Adaptive Decision Framework} (DSADF), integrating two complementary modules: System 1, comprising an RL agent and a memory space for fast and intuitive decision making, and System 2, driven by a VLM for deep and analytical reasoning. DSADF facilitates efficient and adaptive decision-making by combining the strengths of both systems. The empirical study in the video game environment: \textit{Crafter} and \textit{Housekeep} demonstrates the effectiveness of our proposed method, showing significant improvements in decision abilities for both unseen and known tasks.
}




\maketitle

\section{Introduction}
\par The reinforcement learning (RL) framework, which is centered on maximizing cumulative rewards, involves an agent iteratively interacting with an environment to explore and refine its decision-making process, ultimately working towards a definite goal. RL has demonstrated remarkable success in a range of applications, including robotics \cite{ibarz2021train, peng2018sim}, video games \cite{mnih2013playing}, dialogue systems \cite{du2019provably}, and autonomous vehicles \cite{chen2019model}. 
\par The critical challenge of RL agents is their generalizability to new unseen settings \cite{di2022goal,yang2019generalized}. This limitation arises because agents rely on experiences specific to the training environment. More concerning is that even minor environmental modifications can substantially affect the agent’s policy generalization ability \cite{ao2023curriculum, rank2024performative}. The open environments to which agents must adapt are usually dynamically changing, making the learning process more complex. Agents must adapt to environmental changes, which requires the proposal of advanced RL methods.
\textcolor{black}{In addition, the RL process remains consistently challenging \cite{cao2024survey, dalal2024plan} for the long-horizon tasks. One critical issue is the sparse rewards caused by the Credit Assignment Problem (CAP) \cite{cao2024survey}, where the agent struggles to correctly attribute the final reward to key decisions in the sequence, leading to reduced learning efficiency and difficulty optimizing its policy. Simultaneously, RL agents should learn high-level reasoning and low-level control \cite{dalal2024plan}. This dual learning mechanism increases the complexity of training, as high-level reasoning requires the agent to plan long-term strategies, while low-level control involves fine-grained, immediate decisions.}
\par Recent research has explored leveraging the prior knowledge and natural language understanding capabilities of foundation models to improve the generalization of RL agents. These foundation models have been used for the extraction of multimodal features, integrating information from various modalities such as vision and language, enabling RL agents to understand complex environments~\cite{stooke2021decoupling, laskin2020curl, schwarzer2020data}. In addition, Large Language Models (LLMs) and Vision Language Models (VLMs) can serve as decision-making assistants, using their reasoning and planning capabilities to support RL agents in formulating actions and strategies~\cite{chakraborty2023re, majumdar2020improving}. Through mechanisms like the Chain of Thought (CoT), LLMs and VLMs facilitate complex reasoning in RL agents, allowing them to decompose difficult tasks and adopt step-by-step problem solving approaches~\cite{du2023guiding, yan2023ask, li2024auto}. In some cases, the foundation models are independent agents, executing entire task processes autonomously~\cite {inoue2022prompter, li2022pre}. Nevertheless, despite the valuable insights of these works, they suffer from an unideal collaboration between RL agents and foundation models, leading to under-utilization of foundation models' reasoning capabilities, inefficient execution and interaction. Moreover, some of these efforts do not address the challenges CAP poses to RL, causing difficulty in optimizing its policies \cite{cao2024survey}.
\par The key challenge of this problem is to design \textcolor{black}{a synergic interaction mechanism that integrates the knowledge of the RL agent in navigating specific environments (exceptional in short-horizon tasks) with the knowledge and reasoning skills of the foundation models (delivering rational responses even for unknown cases). }Positioning LLMs and VLMs as rational long-term planning systems and RL agents as intuitive short-term reactive systems, while facilitating interaction between the two, is a research assumption that has not been adequately explored.
%
\par
\begin{figure}[!t]
    \centering
    \includegraphics[width=10cm]{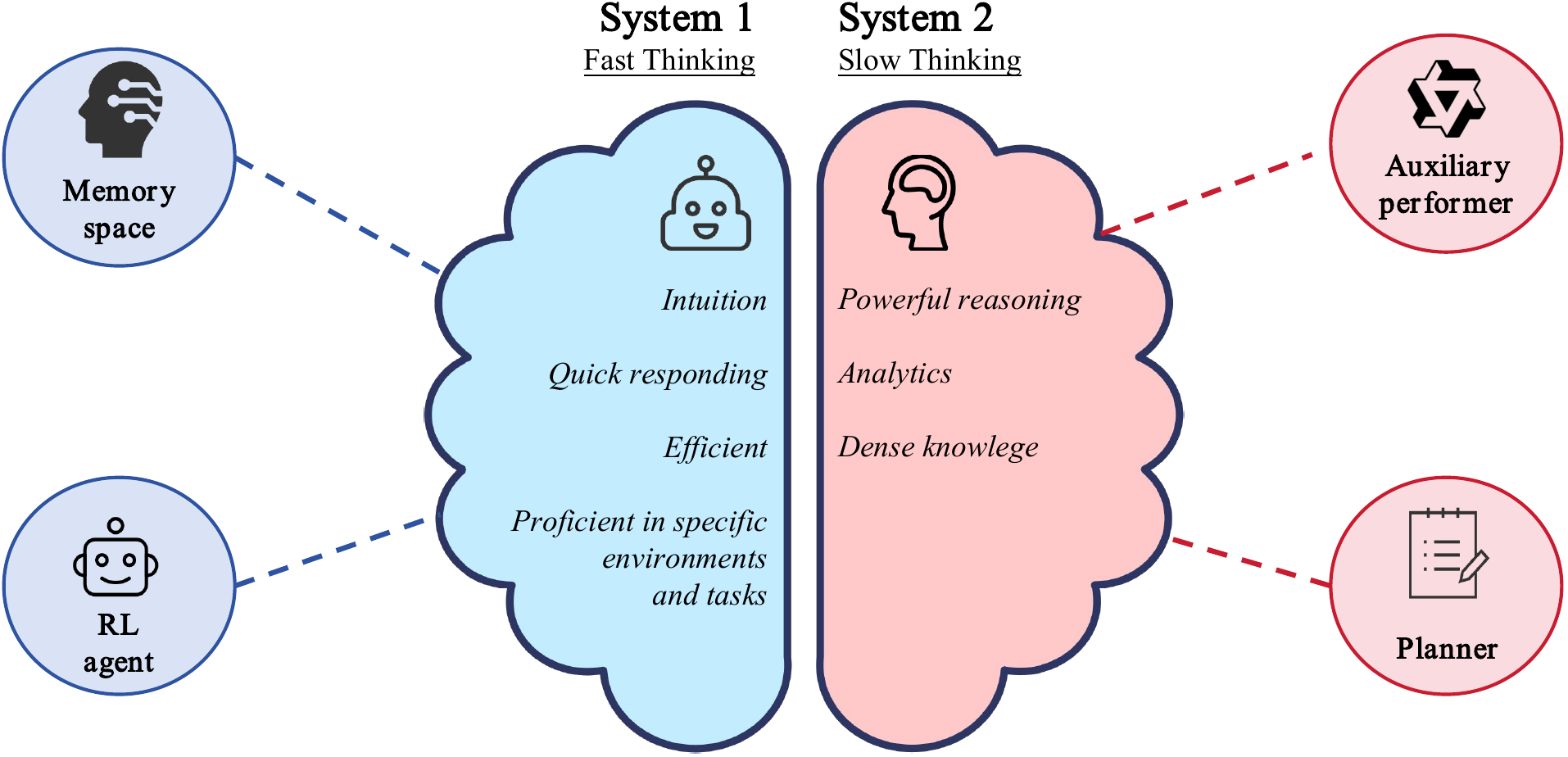} 
    \caption{Fast thinking and slow thinking}
    \label{fig:dual_system}
\end{figure}

 \textcolor{black}{To this end, we draw inspiration from Kahneman’s cognitive theory \cite{daniel2017thinking}, particularly the dual-process framework of ``System 1" and ``System 2," which enhances both the efficiency of the decision and generalization. Fig. \ref{fig:dual_system} illustrates such an insight that the planning ability of System 2 can boost the decision-making pipeline of System 1.}
Essentially, two modes of human thinking are distinguished: System 1, a fast, intuitive process driven by emotions, memory, and experience, and System 2, a slower, more deliberate approach requiring focused attention for analytical problem-solving \cite{wei2022chain, jin2024graph, zhou2024analyzing}. 
The dual-system approach has been preliminarily implemented in areas such as Visual Question Answering (VQA) \cite{sun2024visual} and Continual Learning (CL) \cite{qi2024interactive}, demonstrating improvements in both efficiency and generalization. However, to the best of our knowledge, no research has explored the application of dual-system concepts in the more challenging decision-making of RL agents.
\par
\begin{figure*}[!t]
    \centering
    \includegraphics[width=10cm]{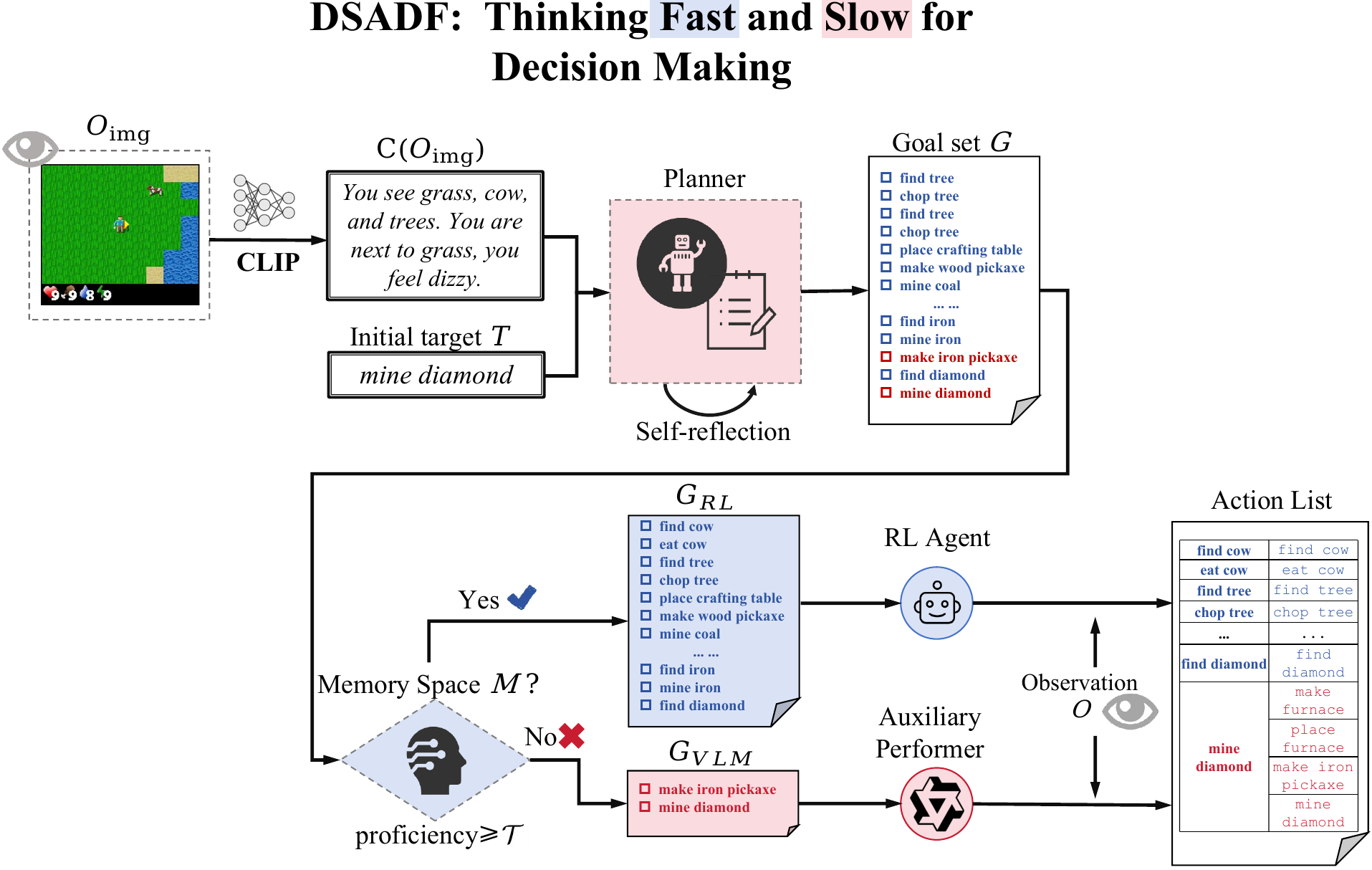}
    \caption{Flowchart of DSADF inference. The \colorbox{mybluegray}{\strut \textbf{blue}} region highlights the fast system components, in contrast to the \colorbox{mypink}{\strut \textbf{pink}} region, which indicates the slow system components.}
    \label{fig:enter-label}
\end{figure*}

 Building upon such a motivation, we propose the \textbf{D}ual-\textbf{S}ystem \textbf{A}daptive \textbf{D}ecision \textbf{F}ramework (DSADF), an innovative approach with an efficient collaboration mechanism integrating System 1 and System 2 to achieve the \textcolor{black}{generalized} decision-making. In this framework, System 1 comprises an RL agent and a memory module, whereas System 2 consists of a VLM. \textcolor{black}{System 2 harnesses the powerful reasoning capabilities of the foundation models to extract clues from instruction hints, breaking down long-horizon initial tasks into manageable single-step tasks. It enables the RL agent to concentrate on solving each single-step task individually, significantly boosting its learning and execution efficiency and maximizing its strengths in single-step tasks. Meanwhile, the VLM continuously updates the memory module based on historical steps, assessing the proficiency of RL agents with different single-step tasks. This process allows for more precise task allocation during testing, ensuring each task is assigned to the suitable performer. Simultaneously, the VLM can leverage its understanding and reasoning capabilities to generate goal-oriented instructions based on goals and observations, facilitating the proficiency boosting of the RL agent.} Our proposed DSADF is illustrated in Figure \ref{fig:enter-label}. The DSADF pipeline starts with System 2, where the VLM serves as a Planner. 
\par We evaluate the effectiveness of DSADF in two video game simulation environments with long-horizon tasks and unseen tasks for the RL agent, \textit{Crafter} and \textit{Housekeep}. In our empirical study, System 1 and System 2 work together synergistically and seamlessly, demonstrating outstanding performance in efficiency and generalization.
\par Our key contributions are as follows:
\begin{itemize}
    \item We propose the \textbf{D}ual-\textbf{S}ystem \textbf{A}daptive \textbf{D}ecision \textbf{F}ramework (DSADF), which draws on the principles of System 1 and System 2. Both systems collaborate so that the proposed DSADF dynamically adapts to long-horizon tasks with high efficiency and good performance.
    \item DSADF shows strong generalization on the unseen tasks of RL agent, demonstrating the impressive performance in various complex environments such as video game \textit{Crafter} and \textit{Housekeep}.   
\end{itemize}

\section{Related Work}
\subsection{Reinforcement Learning with Foundation Models}
\par In recent years, LLMs and VLMs have shown remarkable potential in enhancing the generalization capabilities in RL \cite{chakraborty2023re,pang2024natural,choi2023efficient,pang2023natural,szot2023large,zala2024envgen,carta2023grounding}. To enhance generalization, Pang et al.~\cite{pang2024natural} proposed an RL framework that translates texts into simplified task instructions, allowing agents to understand and perform tasks more efficiently by leveraging LLMs. Carta et al.~\cite{carta2023grounding} enabled LLMs to achieve semantic grounding in interactive environments through online RL, enhancing their task-solving capabilities and sample efficiency. Chakraborty et al.~\cite{chakraborty2023re} introduced a method that applies language feedback to refine RL agent policies, facilitating improved generalization of RL agents in unfamiliar environments. Furthermore, the state-of-the-art foundation models can act as planners \cite{zhou2024generalizable}, decision-makers \cite{du2023guiding, yan2023ask}, and information processors \cite{coleman2024llm} across various scenarios and tasks, enhancing the planning performance of RL agents.
\par LLMs and VLMs can also assist RL agents in handling long-horizon tasks \cite{zhang2024can,janner2021offline,song2023llm,dalal2024plan,zhou2024generalizable,du2023guiding}. These corresponding studies take advantage of the powerful common sense reasoning, dynamic adaptation, and modular hierarchical planning structure of foundation models, applying them to video games \cite{du2023guiding}, robotics \cite{dalal2024plan}, and embodied agents \cite{zhou2024generalizable}.

\subsection{Reasoning of Foundation Models}
In the past few years, LLMs and VLMs have shown astonishing performance, giving people a glimpse of the dawn of general artificial intelligence~\cite{chang2024survey, kasneci2023chatgpt,bubeck2023sparks}. These models, built on the Transformer architecture, leverage advanced technologies to excel in processing and generating natural language. With hundreds of billions or even trillions of parameters, LLMs and VLMs demonstrated an emerging capability for advanced reasoning \cite{zhai2024enhancing,lightman2023let,qiao2022reasoning}. These models can emulate the step-by-step reasoning process resembled by humans \cite{hao2024llm}, showcasing their potential for complex cognitive tasks. This reasoning mechanism is renowned as CoT~\cite{wei2022chain}. This systematic approach to problem-solving is a hallmark of many fields, including mathematical word problems \cite{kojima2022large,wang2022self,lightman2023let,cobbe2021training}, logical reasoning \cite{yao2023beyond,yao2023thinking,besta2024graph}, and multimodal reasoning \cite{chen2023genome,you2023idealgpt,wu2024v}. In the domain of explainable AI, this systematic methodology is simulated by models that generate step-by-step textual explanations of reasoning and decision-making processes \cite{jacovi2020towards,hua2022system}, which is beneficial to the robustness and transparency of AI systems.

\subsection{Dual System}
In human cognition research, Daniel Kahneman's System 1 (fast thinking) and System 2 (slow thinking) theory \cite{daniel2017thinking} provides profound insights into the dual-process model of human thinking and decision-making. In the field of artificial intelligence (AI), this theory has been adopted by scholars to inform the design and analysis of intelligent systems, offering valuable perspectives for research across diverse domains \cite{sun2024visual,qi2024interactive,christakopoulou2024agents,liu2024robomamba,hagendorff2023thinking,miech2021thinking}.

Qi et al. \cite{qi2024interactive} proposed an interactive continual learning framework that combines System 1 and System 2 models to achieve efficient knowledge acquisition and complex reasoning through collaboration.
Hagendorf et al. \cite{hagendorff2023thinking} demonstrated the effectiveness of fast and slow thinking mechanisms in LLMs, revealing how they combine intuitive decision-making with logical reasoning when tackling complex tasks. The study shows that these models can enhance their reasoning abilities through chain-of-thought processes while exhibiting human-like cognitive traits and potential for advancement.
Liu et al. \cite{liu2024robomamba} proposed a highly efficient Vision-Language-Action (VLA) model called RoboMamba, which combines a Fast pathway for rapid vision-language alignment with a Slow pathway for deep reasoning. This model achieves efficient robotic reasoning and action capabilities while addressing simple tasks and complex scenarios at a low cost. 
Furthermore, System 1 and System 2 principles have proven highly effective across diverse domains, including visual Q\&A \cite{sun2024visual} and information retrieval \cite{miech2021thinking}.
In this study, we introduce a novel decision-making system that seamlessly integrates the strengths of System 1 and System 2. System 1 enables rapid, intuitive processing, allowing for quick responses, while System 2 excels in deep reasoning and complex problem-solving, ensuring generalization and robustness. By combining these complementary cognitive mechanisms, our system achieves both speed and reliability, effectively balancing efficiency with thoughtful decision-making. It not only handles complex tasks with precision and efficiency but also demonstrates strong generalization capabilities, successfully adapting to unseen tasks. This innovative approach pushes the boundaries of traditional decision-making systems, offering broad applications across various domains.

\section{Preliminaries}
\par This section illustrates the preliminaries of this study, including the definition of the mathematic symbols and the problem description.
\subsection{Definition}
\par Let \(\pi(\cdot)\) be the RL agent, which governs the action selection process based on given tasks. The Memory Space is defined as \(\mathcal{M} = \{\mathcal{U}_i\}_{i=1}^K\), where the symbol $\mathcal{U}_i =\{({u_i^n}, p_i^n)\}_{n=1}^{D}$ represents an short-horizon task subspace containing \(D\) short-horizon tasks \(u_i^n\) of the type \(i\).
The CoT of the vision language model, \(VLM(\cdot)\), decomposes the goal tasks into simpler components or classifies the unseen objects. The observation tuple is defined as \((O_{\text{img}}, O_{\text{text}})\). In this tuple, \(O_{\text{img}}\) represents the visual observation of the environment, and \(O_{\text{text}}\) denotes the corresponding textual observation. The textual observation \(O_{\text{text}}\) corresponds to the visual observation \(O_{\text{img}}\) through the Contrastive Language-Image Pre-training (CLIP) module~\cite{CLIP} with inference \(C(\cdot)\). The knowledge conveyed by \(O_{\text{text}}\) includes the object coordinates from the goal list \(G\) and the corresponding action subspace of the object.
Subsequently, the knowledge conveyed by \(O_{\text{text}}\) includes the object coordinates from the goal list \(G\) and the corresponding action subspace of the object.
The initial target \(T\) is the overall goal and can be decomposed by the VLM into step-by-step goals \(G\) using textual observation description $C(O_{\text{img}})$, which can be derived as \(G = VLM(T, O_{\text{text}})\). The goal executed by the RL agent is \(G_{\text{RL}}\), and the goal executed by the VLM is \(G_{\text{VLM}}\). Together, they form the complete set of goals \(G = G_{\text{VLM}} \cup G_{\text{RL}}\). The goal list $G_{\text{RL}}$ is a guiding reference for the RL agent, directing System 1’s actions toward achieving the desired outcomes. \(R_\mathrm{ini}\) denotes the initial target reward for the primary objective, \(R_{\text{sub}}\) represents intermediate rewards for achieving sub-goals, and \(R_{\text{proxi}}\) incentivizes the agent to maintain proximity to target objects or regions during task execution. Besides, we summarize important notations in Table \ref{tab_notation}.
\begin{table}[!t]
\tabcolsep 1cm
  \centering 
  \caption{Notations and Definitions}
    \begin{tabular}{c|c}
    \midrule[1.2pt]
    Notation & Definition \\
    \midrule[1.2pt]
    $VLM(\cdot)$   & Vision language model's reasoning \\
    $O_{\text{img}}$   & Image observation \\
    $O_{\text{text}}$   & Textual observation \\
    $C(\cdot)$   & CLIP network's inference \\
    $T$   & Initial target \\
    $G$   & Goal list \\
     $ G_{}$ &  Initial goal set\\
    $G_{\text{RL}}$ & RL agent's Goal set \\
    $G_{\text{VLM}}$ & VLM agent's Goal set \\
    $\pi(\cdot)$   & RL agent \\
    $M $  & Memory Space \\
    $H$   & Hint instruction  \\
    $\mathcal{U}$   & Short-horizon task subspace \\
    $e_d$ & reflection instruction \\
    \(R_\mathrm{ini}\) & Initial target reward  \\
    $R_{\text{sub}}$ & Sub-goal Reward  \\
    $R_{\text{proxi}}$ & Proximity reward \\
    $E_{\text{text}}(\cdot)$ & Text encoder \\
     $E_{\text{img}}(\cdot)$ & Image encoder \\
    \midrule[1.2pt]
    \end{tabular}
  \label{tab_notation}
\end{table}
%
%

\subsection{Problem Description}
\label{goals_set}
\par In conventional RL, task execution is often modeled as a Partially Observable Markov Decision Process (POMDP), represented by the tuple $(S, A, \pi, T, O, R, \gamma)$. Here, $S$ denotes the state space encompassing the agent and its environment, and $A$ represents the set of actions available to the agent. The policy network is defined by $\pi$, and $T$ describes the transition dynamics, specifying the probability of moving between states given a particular action. The observation function $O$ maps states and actions to observations, and the discount factor $\gamma$ controls the weighting of future rewards.
\par The history at time step $t$, denoted as $h_t$, consists of the sequence of observations and actions taken up to time $t-1$, denoted as $h_t = (o_0, a_0, o_1, a_1, \dots, o_{t-1}, a_{t-1})$. For goal-conditioned RL \cite{du2023guiding}, the optimal policy \( \pi(h_t|g) \), which maximizes the expected cumulative reward, can be expressed as follows:
\begin{equation}
\pi(h_t|g) = \arg \max_{a \in A} \mathbb{E} \left[ \sum_{i=0}^{\infty} \gamma^i R(s_{t+i}, a_{t+i}\mid a_t = a,g) \right],
\label{eq1_reward}
\end{equation}
where $g$ represents a specific goal for the RL agent. 
\par However, two key challenges persist for goal-conditioned RL agents: (1) The agent struggles to achieve goals \(g\) that are not incorporated into the training simulations, and (2) multi-step goals \(g\) introduce a combinatorial explosion in the action space, significantly reducing the agent's generalization ability. To address these problems, we propose a DSADF method based on the cognitive theory, and its decision output can be denoted as $\pi_{\text{DSADF}}(\cdot)$. 
\par The main goal of DSADF is to design a reliable decision-making system based on an RL agent and a VLM. The decision-making mechanism we develop should meet the following objectives:
\begin{itemize}
\item {\bf Object 1: Accurately complete tasks.} One of the key objectives of our system is to precisely achieve the desired goal \(g\). This requires the system to consistently interpret task requirements and take actions that effectively drive toward the successful completion of the objective.
\item {\bf Object 2: Accomplish tasks efficiently.} Efficiency can be demonstrated in two ways: 1) completing more tasks with fewer testing action steps and 2) achieving better performance with fewer training steps. 
\item {\bf Object 3: Powerful generalization ability.} Our goal is for DSADF to make informed decisions and complete tasks by leveraging powerful reasoning and extensive background knowledge, even when encountering unfamiliar items, scenarios, or tasks during the training process.
\end{itemize}


\section{Method}
\par Our Dual-System Adaptive Decision Framework (DSADF) is composed of two key components: fast thinking (System 1) and slow thinking (System 2). System 1 is the primary performer, delivering quick responses to diverse environments and tasks. Meanwhile, System 2 operates as a Planner with deep reasoning capabilities, using its vast knowledge and reasoning skills to supervise System 1 during both the training and testing phases. Furthermore, System 2 acts as an auxiliary performer, interacting with the environment to address states or short-horizon tasks unfamiliar to System 1. System 1 and System 2 achieve high efficiency and precision by working in tandem, excelling in complex environments. Subsection \ref{sys1} introduces System 1, followed by the presentation of System 2 in Section \ref{sys2}. 
\par Our proposed DSADF transforms the long-horizon initial target $T$ into a sequence of short-horizon task sets, denoted as $G$. Based on the RL agent's proficiency with different short-horizon tasks, as stored in the memory space, these short-horizon tasks in $G$ are divided into two subsets: $G_{\text{RL}}$, which consists of tasks handled by the RL agent, and $G_{\text{VLM}}$, assigned to the VLM. The tasks in $G_{\text{RL}}$ and $G_{\text{VLM}}$ will be executed alternately according to the priority order in $G$.
\subsection{Fast Thinking Component (System 1)}
\label{sys1}
\par System 1 is a primary performer, responding swiftly to familiar observations and tasks. It comprises two main components: the goal-conditional RL agent \(\pi(\cdot)\) and the Memory Space \(M\), which collaborate to facilitate rapid recognition and execution in familiar short-horizon tasks.

\par \textbf{Goal-conditional RL agent:} In the DSADF framework, the goal-conditional RL agent is the core of System 1, responding to environmental observations. At each state \(t\), the RL agent is provided with a goal \(g^t \in G_{\text{RL}}^t\) and an observation pair \((O_{\text{img}}^t, O_{\text{text}}^t)\). The agent uses the encoders \(E_{\text{img}}(\cdot)\) and \(E_{\text{text}}(\cdot)\) to obtain embeddings \(E_{\text{img}}(O_{\text{img}}^t)\) and \(E_{\text{text}}(O_{\text{text}}^t)\) for image and text, respectively. Based on these embeddings, along with the goal text embedding \(E_{\text{text}}(g^t)\), it determines the appropriate action \(a^t\) as:
\begin{equation}
a^t = \pi(E_{\text{img}}(O_{\text{img}}^t), E_{\text{text}}(O_{\text{text}}^t), E_{\text{text}}(g^t)).
\end{equation}
When objects are present in both the goal list \(G_{\text{RL}}^t\) and the observation text \(O_{\text{text}}^t\), the observation text also includes their respective coordinates, providing directional and distance information for effective navigation and task execution.

\par \textbf{Memory Space:} \textcolor{black}{The Memory Space \(M = \{\mathcal{U}_i\}_{i=1}^{K}\) consists of \(K\) short-horizon task subspaces, each representing a category of short-horizon tasks with shared actions. For instance, the ``attack" subspace can include a series of short-horizon tasks related to attack actions, such as ``attack cow" and ``attack skeleton." 
After training the RL agent, we record all the actions that received rewards and store them in the subspace \( \mathcal{U} \) for further evaluation of the agent's proficiency in these tasks. Each subspace \( \mathcal{U} \) contains pairs of short-horizon tasks and their corresponding proficiency values, denoted as \( \mathcal{U} = \{(u^v, p^v)\}_{v=1}^V \) and $V$ represents the numbers of rewarded actions. Each action \( u^v \in \mathcal{U} \) is tested individually.
For each action \( u^v \in \mathcal{U} \), we provide the information \( h_v \) (e.g., action steps, initial situation). The agent will provide the VLM with the state observation for evaluation and obtain the corresponding proficiency value \( p^v \), where \( p^v = \text{VLM}(u^v, h_v) \). A higher proficiency value indicates that the RL agent has consistently and effectively performed the corresponding short-horizon task \(u^v\), demonstrating mastery across various situations. Based on the value of \(p^v\), the system determines the necessity of delegating the RL agent to System 2. Further details on the planning process can be found in Section \ref{sec:Planner}.}

\par \textcolor{black}{This dynamic task allocation ensures that the RL agent’s capabilities on short-horizon tasks are fully utilized, enhancing the system’s overall execution performance. Ultimately, the presence of System 1 not only significantly accelerates response times but also provides a solid foundation for System 2 to handle more complex reasoning and planning tasks.}

\subsection{Slow Thinking Component (System 2)}
\label{sys2}
\par System 2 is the brain of the entire system, endowed with abundant knowledge and advanced reasoning capabilities. Leveraging this intelligence, it primarily acts as the system's Planner, which can solve long-horizon tasks, particularly challenging for RL agents in System 1.
Additionally, System 2 is an auxiliary performer, interacting with the environment to handle tasks where the RL agent is not proficient in such a task.


%


\subsubsection{Planner}
\label{sec:Planner}

The RL agent faces several challenges in handling long-horizon tasks, including the difficulty of exploring high-dimensional continuous action spaces, sparse rewards arising from the CAP \cite{pignatelli2024survey}, and the complexity of simultaneously learning both high-level reasoning and low-level control \cite{dalal2024plan}. For RL agents, converting long-horizon tasks into single-step tasks is essential in both the training and testing phases. However, effectively decomposing tasks into sub-sequences and executing each accurately remains a significant challenge \cite{sutton1999between,parr1997reinforcement}.


Leveraging the extensive background knowledge and planning capabilities, VLM works as the Planner to transform the long-horizon initial target \( T \) into a short-horizon initial goal set \( G_{\text{init}} \), based on hint instruction $H$ and the textual description of the observation \( C(O_\mathrm{img}) \). This process can be expressed as:
\begin{equation}
    G_{\text{init}} = VLM(T,C(O_\mathrm{img}),H).
    \label{transformation}
\end{equation}
Each element $g \in G_{\text{init}}$ represents a short-horizon task, ensuring that complex objectives are broken down into manageable actions.
In the planning process, we used Chain-of-Thought (CoT) \cite{wei2022chain} to facilitate a more nuanced and comprehensive analysis. The initial question, $q$, is meticulously deconstructed into a series of subset $G_{\text{init}}$. Moreover, we integrate a self-reflection mechanism into our planning process, which serves as a reflective checkpoint to critically evaluate the plans we generate.
\textcolor{black}{As shown in Eq. (\ref{valation}), we derive the reflection instruction $e_t$ from the initial target $T$ and the initial subset \( G_{\text{init}} \). Subsequently, using the initial goal set $G_{\text{init}}$ and the reflection instruction $e_d$ as Eq. (\ref{valation}), we construct a complete goal set \( G \) as follows:}

\begin{equation}
    G = VLM(e_d,G_{\text{init}}).
    \label{final_goal}
\end{equation}

Self-reflection ensures the accuracy of our plans by systematically verifying each component within the subset \( G \). By rigorously examining every step, DSADF can identify and address potential inconsistencies or errors, enhancing the overall reliability and precision of the planning outcomes. 

\par Each sub-task \( g \in G \) is first checked in the Memory Space \( M \) to determine if there is a corresponding short-horizon task \( u \). If such a task \( u \) exists within \( M \), \( g \) is categorized according to Eq. (\ref{assign}) into one of two groups: \( G_{\text{VLM}} \) or \( G_{\text{RL}} \). If no corresponding \( u \) is found in \( M \), \( g \) is directly assigned to \( G_{\text{VLM}} \). Eq. (\ref{assign}) denotes this formulation:
 \begin{equation}
  \begin{cases} 
g \in G_{\text{VLM}} & \text{if } p < \mathcal{T} , \\
g \in G_{\text{RL}} & \text{if } p \geq \mathcal{T}, 
\end{cases}
\label{assign}
\end{equation}
where $p$ is the proficiency value corresponding to \( u \), and $\mathcal{T}$ is a hyper-parameters. The RL agent executes the goals in $G_{\text{RL}}$, and the VLM acts as an auxiliary performer to handle the goals in $G_{\text{VLM}}$, as explained in Section~\ref{performer}. The two systems collaborate to handle the out-of-distribution tasks during the testing phase.

\noindent \textbf{Self-reflection:} \textcolor{black}{For the initially generated task \( G_{\text{init}} \), we will perform a reflection and validation process, which will yield a reflection instruction \( e_d \). The process can be described as:} 
 \begin{equation}
  e_d = VLM(G_{\text{init}},T),
\label{valation}
\end{equation}
\textcolor{black}{where $T$ denotes the initial target. Based on this reflection instruction, we will then regenerate the task list \( G \) as Eq. (\ref{final_goal}).}

\begin{algorithm}[t!]
    \caption{DSADF training process}
    \label{algo:rl_training}
    \begin{algorithmic}[1]
        \renewcommand{\algorithmicrequire}{\textbf{Input:}}
        \renewcommand{\algorithmicensure}{\textbf{Output:}}
        \Require Long-horizon task $T$, observation space $(O_{\text{img}}, O_{\text{text}})$, VLM Planner $VLM(\cdot)$, proficiency threshold $\mathcal{T}$, Clip model $C(\cdot)$, instruction hint $H$. 
        \Ensure Trained RL policy agent $\pi(\cdot)$, memory space $M$.
        
        \State Initialize RL policy $\pi$, memory space $M \gets \emptyset$
        \State \textcolor{blue}{//Task Decomposition by Planner}
        \For{each episode d}
        \State Generate initial goals: $G_{\text{init}} \gets VLM(T, C(O_{\text{img}}), H)$ \label{line:decomp}
        \State Self-reflection: $e_d \gets VLM(G_{\text{init}}, T)$ \label{line:reflect}
        \State Finalize goals: $G \gets VLM(e_d, G_{\text{init}})$ \label{line:finalgoals}
        
        \State \textcolor{blue}{// Progressive Reward Training}
        \For{each step $t$}
        \State Get current goal $g^t \in G$
        \State Encode inputs: 
        \State \quad $E_{\text{img}} \gets E_{\text{img}}(O_{\text{img}}^t)$, $E_{\text{text}} \gets E_{\text{text}}(O_{\text{text}}^t)$
        \State \quad $E_g \gets E_{\text{text}}(g^t)$
        \State Select action: $a^t \gets \pi(E_{\text{img}}, E_{\text{text}}, E_g)$ \label{line:action}
        
        \State \textcolor{blue}{// Compute Hierarchical Rewards}
        \State Sparse reward: $R_{T} \gets  \text{sgn}(\delta(T))$
        \State Sub-goal reward: $R_{\text{sub}} \gets \sum_{j=1}^J \text{sgn}(\delta(g_{s_j}))$ for $g_{s_j} \in G_{\text{sub}}$
        \State Proximity reward: Computing $R_{\text{proxi}}$ via Eq. (\ref{eq_pro}) with action  $a^t 
        $
        \State Total reward: $R \gets \gamma_1 R_{\text{T}} + \gamma_2 R_{\text{sub}} + \gamma_3 R_{\text{proxi}}$ \label{line:reward}
        
        \State Update $\pi$ using $R$ 
        \EndFor
        \EndFor
        \State Update memory $M$ with new proficiency $p^v \gets VLM(u^v, h_d)$ for action $u^v$ based evaluation \label{line:memory}
    \end{algorithmic}
\end{algorithm}

\subsubsection{Auxiliary performer}
\label{performer}
\par In section \ref{sec:Planner}, we divide the decomposed goal list \( G \) into the RL agent goal set \( G_{\text{RL}} \) and the VLM goal set \( G_{\text{VLM}} \). When the RL agent faces unfamiliar tasks, the VLM \( G_{\text{VLM}} \) works as an auxiliary performer to handle such a scenario.
With extensive pre-training, VLMs possess a broad knowledge base, advanced natural language understanding and generation capabilities, multi-modal information processing abilities, and strong generalization skills \cite{vafa2024large,saba2024llms,baai2023plan4mc}.
These attributes enable VLMs to generate goal-related instructions based on goals and observations, allowing them to function as independent agents to accomplish goals \cite{zhou2024generalizable}. 
Therefore, the VLM can act as an auxiliary performer, leveraging its strong generalization abilities to tackle tasks beyond the RL agent’s expertise, creating an efficient collaborative mechanism. Additionally, the VLM can handle real-time economic tasks, further enhancing its role as an auxiliary performer.

\noindent\textbf{Executing Tasks in \( G_{\text{VLM}} \):} When a goal belongs to \( G_{\text{VLM}} \), the VLM acts as an auxiliary performer, handling tasks unfamiliar to the RL agent, defined as \( G_{\text{VLM}} \in G_{\text{VLM}} \). In this case, the VLM generates the corresponding goal-related instruction \( t_\mathrm{VLM} \) based on the textual observation \( O_{\text{text}} \) and the goal \( G_{\text{VLM}} \), expressed as \( t_\mathrm{VLM} = VLM(O_{\text{text}}, G_{\text{VLM}}) \). The goal-related instruction \( t_\mathrm{VLM} \) can interact with the environment to achieve the goal.

\noindent\textbf{Handling unseen situation:}
When an object \( o \in O_{\text{text}} \) is absent from the Memory Space $M$ and is positioned near our agent, the VLM will function as an auxiliary performer, issuing temporary emergency instructions $t$.

\noindent \textbf{Implementation:} \textcolor{black}{When VLM operates as an Auxiliary Performer, it is responsible for carrying out the plan defined by the planner. In practice, we employ a smaller VLM (e.g., Qwen 2.5- VL 7B) as the performer for each action, reducing token costs and ensuring faster response times.}



\begin{algorithm}[t!]
    \caption{DSADF Inference Execution}
    \label{algo:inference}
    \begin{algorithmic}[1]
        \renewcommand{\algorithmicrequire}{\textbf{Input:}}
        \renewcommand{\algorithmicensure}{\textbf{Output:}}
        \Require Initial target $T$, observation $(O_{\text{img}}, O_{\text{text}})$, \quad Memory space $M$, RL policy $\pi$, VLM Planner/Performer, proficiency values $p^v \in M$, Threshold value $\mathcal{T}$
        \Ensure Completed task set $\mathcal{T}_{\text{comp}}$
        
        \State $\mathcal{T}_{\text{comp}} \gets \emptyset$
        \State \textcolor{blue}{// Slow Thinking (System 2) for Task Decomposition \& Routing}
        \State Generate initial goals: $G_{\text{init}} \gets VLM(T, C(O_{\text{img}}), H)$ \label{line:decomp}
        \State Self-reflection: $e_d \gets VLM(G_{\text{init}}, T)$ \label{line:reflect}
        \State Finalize goals: $G \gets VLM(e_d, G_{\text{init}})$ \label{line:finalgoals}
        \State Partition $G$ into $G_{\text{RL}}$, $G_{\text{VLM}}$ via:
        \For{each $g \in G$}
            \If{$\exists (u^v, p^v) \in M \text{ where } u^v \equiv g$}
                \State Assign $g$ to $G_{\text{RL}}$ if $p^v \geq \mathcal{T}$ else $G_{\text{VLM}}$ \label{line:routing}
            \Else
                \State $G_{\text{VLM}} \gets G_{\text{VLM}} \cup \{g\}$ \label{line:novel_task}
            \EndIf
        \EndFor
        
        \State \textcolor{blue}{//Alternating Execution}
        \While{$G_{\text{RL}} \neq \emptyset \lor G_{\text{VLM}} \neq \emptyset$}
            \State Get highest priority goal $g^t$ from $G$ \label{line:priority}
            \If{$g^t \in G_{\text{RL}}$} 
                \State $a^t \gets \pi(E_{\text{img}}(O_{\text{img}}^t), E_{\text{text}}(O_{\text{text}}^t), E_{\text{text}}(g^t))$ \textcolor{blue}{// Fast Thinking (System 1) for Action Execution \& Routing}\label{line:rl_act}
                \State Execute $a^t$, receive new observation $o^{t+1}$
                \If{$\delta(g^t) = 1$} \Comment{Check completion}
                    \State $\mathcal{T}_{\text{comp}} \gets \mathcal{T}_{\text{comp}} \cup \{g^t\}$
                    \State Remove $g^t$ from $G_{\text{RL}}$
                \EndIf
            \Else
                \State $t_{\text{vlm}} \gets \text{VLM}(O_{\text{text}}^t, g^t)$ \label{line:vlm_act}
                \State Execute $t_{\text{vlm}}$, update $o^{t+1}$
                \If{$\delta(g^t) = 1$}
                    \State $\mathcal{T}_{\text{comp}} \gets \mathcal{T}_{\text{comp}} \cup \{g^t\}$
                    \State Remove $g^t$ from $G_{\text{VLM}}$
                \EndIf
            \EndIf
            
            \State \textcolor{blue}{// Emergency Handling based on auxiliary performer}
            \If{$\exists o \in O_{\text{text}}^{t+1} \notin M \text{ near agent}$}
                \State Generate emergency $t_{\text{emerg}} \gets \text{VLM}(o)$
                \State Execute $t_{\text{emerg}}$ immediately \label{line:emergency}
            \EndIf
        \EndWhile
    \end{algorithmic}
\end{algorithm}

\subsection{Progressive  Reward for RL Agent}
\label{reward}
\par The goal-conditioned progressive reward is denoted in Eq. (\ref{eq1_reward}).
Our reward function includes three components. The initial target reward \(R_\mathrm{ini}\) determines whether the initial target \( T \) has been achieved. Moreover, the sub-goal reward $R_\mathrm{sub}$ is based on the completion of decomposed sub-goals $g \in G$. Last but not least, the proximity reward $R_\mathrm{proxi}$ guides the agent closer to each sub-goal $g$. This hierarchical structure helps provide a guidance for the RL agent's behavior.
The following paragraphs provide an essential interpretation of these three types of rewards.

\textbf{Initial target reward \(R_\mathrm{T}\)}: This is a sparse reward, meaning the reward is only granted upon completion of each task within the initial target \( T \). Feedback is provided when a specific component of \( T \) is accomplished as follows:
\begin{equation}
R_{T} =  \operatorname{sgn}(\delta(T)),
\end{equation}
where the function \(\operatorname{sgn}(\cdot)\) represents the sign function, and \(\delta(\cdot)\) denotes the indicator function. Specifically, \(\delta(T)\) is triggered only when the initial task \(T\) is successfully completed.

\textbf{Sub-goal reward $R_{\text{sub}}:$} This is a dense reward function. The sub-goals \( G_{\text{sub}} \) are defined as the complement of the initial target \( T \) within the complete goal set \( G \) generated in Eq. (\ref{transformation}), represented as \( G_{\text{subgoal}} = G \cap T^c \). These sub-goals are inferred by the VLM based on hint instructions $H$ and act as essential prerequisite tasks for completing the initial target \( T \). It means that to reach the final target \( T \), the model must progressively achieve these decomposed sub-tasks \( G_{\text{subgoal}} \). Establishing these sub-goals breaks down the complex task \( T \) into smaller, manageable parts, enabling a step-by-step progression toward the overall objective. The formulation can be shown as:
\begin{equation}
R_{\text{sub}} = \sum_{j=1}^{J} \operatorname{sgn}(\delta(g_{s_j})), \quad \text{where } g_{s_j} \in G_{\text{subgoal}}.
\end{equation}

\textbf{Proximity reward $R_{\text{proxi}}$}: Building on the bags of tricks in previous RL studies \cite{fan2022minedojo, du2023guiding}, we design a proximity reward to provide a denser reward signal, allowing the RL agent to learn each sub-goal \( g \in G \) more efficiently. This proximity reward encourages actions beneficial to achieve the sub-goal, thereby narrowing the exploration space of possible actions. Inspired by the previous study \cite{du2023guiding}, we calculate the cosine similarity between the embedding of each sub-goal $E_g$ and the agent’s transition in the environment $E_{\mathrm{trans}}$ as a proximity reward to guide the agent's actions towards achieving each sub-goal. The representation can be shown as:
\begin{equation}
\resizebox{0.8\linewidth}{!}{
$
R_{\text{proxi}}(o, a, o' | g) = 
\begin{cases} 
\text{cos}(E_\mathrm{trans}, E_g) & \text{if } \text{cos}(E_\mathrm{tran}, E_g) > \beta  \\
0 & \text{otherwise}
\end{cases}
$
}
,
\label{eq_pro}
\end{equation}
where \( o \) is the observation, and \( o' \) is the updated observation after taking action \( a \). \( E_g = E_{\text{text}}(g), g \in G \), and \( E_{\mathrm{trans}} = E_{\text{text}}(C_{\mathrm{trans}}(o, a,o') \), where \( C_{\mathrm{trans}}(o, a,o') \) represents the description of the agent’s transition in the environment. $\text{cos}(\cdot, \cdot)$ is the cosine similarity function, and we use a pre-trained SentenceBERT \cite{reimers2019sentence} as text encoder $E_{\text{text}}(\cdot)$ to obtain both embeddings. 
When the transition's caption closely matches the sub-goal (i.e., when \(\text{cos}(E_{\mathrm{trans}}, E_g) > \beta\), where \(\beta\) is a predefined similarity threshold), the agent is rewarded in proportion to the degree of similarity.


The final reward \(R\) is computed as the cumulative sum of the three-layered reward components: the initial target reward \(R_{\text{ini}}\), the sub-goal reward \(R_{\text{sub}}\), and the proximity reward \(R_{\text{proxi}}\), each scaled by their respective hyperparameters \(\gamma_1, \gamma_2, \gamma_3\). The final reward is expressed as:
\begin{equation}
R = \gamma_1 R_{\text{T}} + \gamma_2 R_{\text{sub}} + \gamma_3 R_{\text{proxi}},
\end{equation}
where \(\gamma_1, \gamma_2, \gamma_3\) are hyperparameters that control the relative contribution of each reward component to the overall reward.
Our progressive reward system ensures that the RL agent transforms the sparse reward of the long-horizon initial task \( T \) into a structured dense, reward for the goal list \( G \). This approach allows the RL agent to focus more effectively on learning single-step tasks, maximizing its strengths and enhancing the efficiency of the RL training process.

\subsection{Overall process of DSADF}

In this chapter, we provide a detailed introduction to the training and test inference process of DASDF based on System 1 and System 2.

\textbf{DASDF training process:} In the DASDF training process, the primary focus is training the RL agent $\pi(\cdot)$. Initially, System 2 is used to decompose the target target $T$, which is then optimized using progressive reward. The trained RL agent then uses the VLM's evaluation capability in system 2 to assess the proficiency value of the rewarded action, thus updating the memory space $M$. The whole process can be described in Algorithm \ref{algo:rl_training}.

\textbf{DASDF testing inference process:} During the DASDF inference process, System 2 initially serves as a planner to decompose the initial target $T$ into a task goal $G$. By consulting the memory space $\mathcal{M}$, the RL agent assesses its familiarity with the subtasks and partitions $G$ into two disjoint subsets: $G_{\text{VLM}}$ and $G_{\text{RL}}$, where $G = G_{\text{VLM}} \cup G_{\text{RL}}$. Subsequently, the RL agent executes $G_{\text{RL}}$ while the VLM operates as an auxiliary performer to process tasks in $G_{\text{VLM}}$. The process can been seen in Algorithm \ref{algo:inference}.

\section{Experiment}
\par This section provides a comprehensive evaluation of methods based on three dimensions: in-domain generalization, efficiency, and out-of-domain generalization. Subsection \ref{task_description} provides an overview of the various tasks, and Sections \ref{crafter} and \ref{house_keep} present our results in the Crafter \cite{hafner2021benchmarking} and HouseKeep \cite{kant2022housekeep} environments, respectively. The scenario in the Crafter environment emphasizes the performance of DSADF on the metrics of in-domain generalization, efficiency, and out-of-domain generalization, whereas the scenario in the HouseKeep environment focuses on its in-domain and out-of-domain generalization.

\subsection{Total experiment setting}

The \textit{Crafter} and \textit{Housekeep} models are updated using a RL algorithm based on Deep Q-Networks (DQN) \cite{mnih2013playing}. The underlying neural network architecture aligns with that of \cite{du2023guiding}, enabling semantic consistency and improved performance during policy learning. By default, the system uses a pre-trained method as \textbf{E}xploring with LLMs (ELLM) \cite{du2023guiding} as the foundational model, providing strong language understanding capabilities for downstream tasks.

For multimodal processing, both the text encoder and image encoder are based on the ImageBind architecture \cite{girdhar2023imagebind}, allowing for unified cross-modal representation learning. This design not only enhances the model's performance on different tasks but also enriches the semantic features used in state modeling throughout the RL process.

\subsection{Task Description}
\label{task_description}
\par Building on the baseline established by \cite{du2023guiding}, we evaluate our DSADF in two challenging environments: (1) Crafter \cite{hafner2021benchmarking}, a sandbox game within the open world, survival and creation genres, and (2) Housekeep \cite{kant2022housekeep}, a robotics environment designed for simulating household object rearrangement tasks. In these environments, common sense knowledge is essential for constraining exploration space, promoting realistic interactions, and enabling efficient object placement by agents.
\par We design 14 tasks in two game environments, Crafter and Housekeep, to evaluate the \textbf{in-domain generalization} and \textbf{out-of-domain generalization} capabilities of the DSADF system:
\begin{enumerate}
    \item \textbf{Evaluating performance on in-domain tasks}: Tasks 1 to 4 are tasks that are encountered during training. These tasks are designed to evaluate the system's performance on tasks learned during training, ensuring that it can reliably reproduce the behaviors acquired throughout the training process.
    \item \textbf{Evaluating generalization to complex long-horizon tasks}: Tasks 5 and 8 are multi-step, out-of-distribution tasks that evaluate the system's \textbf{generalization} ability to handle novel and complex scenarios, requiring multi-step reasoning and strategic planning.
    \item \textbf{Evaluating generalization to simple tasks}: Tasks 9 to 11 are short tasks comprising 2 to 3 steps, designed to evaluate the system's ability to generalize when handling simpler tasks. Since the VLM agent (a 70B-parameter foundation model) serves as an Auxiliary Performer in our method, this evaluation also provides an indirect assessment of our method's effectiveness compared to using the VLM independently as the Auxiliary Performer.
    \item \textbf{Evaluating performance on exploratory rewards}: Tasks 12 to 14 are part of the Housekeep environment and are specifically designed to assess the system's generalization performance in handling a distinct and challenging task. Unlike environments where explicit reward preferences guide behavior, the Housekeep environment requires the system to rely on its ability to infer which actions result in positive outcomes. This inference is based on analyzing the action history and accumulated rewards, making it a test of the system's adaptability and decision-making under uncertain conditions.
\end{enumerate}
This task framework provides a comprehensive and structured evaluation of the performance of DSADF in terms of \textbf{in-domain generalization} on known environments, as well as its \textbf{out-of-domain generalization} to unseen and varying task complexities. We place the specific settings for each task in Table \ref{tab_notation}.

\subsection{Crafter}
\label{crafter}
\par \textbf{Environment description:} Crafter \cite{hafner2021benchmarking} is a 2D version of Minecraft. Like Minecraft, Crafter features a procedurally generated, partially observable world where players can collect and craft various artifacts arranged within an achievement tree.
In the baseline work~\cite{du2023guiding}, actions are automatically selected based on the agent’s environment. Currently, each action is explicitly tied to a verb-noun combination, eliminating the reliance on automatic context-based interpretation. It makes the agent's behavior more controllable and intuitive. 
The achievement tree lists all possible achievements and their respective prerequisites. In this environment, we evaluate our framework by carrying out a series of tasks. The action space can be shown in Appendix \ref{act_sub_app}.

\noindent \textbf{Experimental Implementation:} We pre-train the RL agent in System 1 following the work ELLM \cite{du2023guiding} and Active Pre-Training (APT) \cite{liu2021behavior}. Subsequently, we fine-tune the agent on the following tasks: mining iron, crafting a wooden sword, deforestation, and attacking a cow. The detailed implementation can be found in the appendix. By default, our fine-tuning process includes a total of 2 million iterations, with each rollout comprising 10,000 steps. The evaluation phase comprises 15,000 iteration steps. If the task is successfully completed, the evaluation stops automatically; otherwise, it continues until all iteration steps are completed. For hypermeters $\gamma_1,\gamma_2$ and $\gamma_3$, we choose 1,0.5 and 0.2 respectably.
%
We apply GPT-4o~\cite{bubeck2023sparks} as VLM for System 2, assigning progressive rewards to the subtasks generated by System 2 based on the framework detailed in Section \ref{reward}. The used prompt used is depicted in the Appendix \ref{meta_promp}.

\noindent \textbf{Evaluation Metrics:} Our evaluation metrics consist of two key types:
\begin{enumerate}
   \item \textbf{Task Success Rate (TSR):} The TSR is defined as the ratio of successfully completed tasks to the total number of rounds of execution. It can be expressed as: 

\[
\text{TSR} = \dfrac{N_\mathrm{STE}}{N_\mathrm{TER}},
\]
where the variables $N_\mathrm{STE}$ and $N_\mathrm{TER}$ denote the number of successful task execution and the total number of execution rounds. A higher TSR indicates greater system success in accurately completing tasks, demonstrating its robustness and reliability.   
\item \textbf{Times:} The execution time (Sec.) is another important metric. We record the execution time for each test and calculate the average execution time across all evaluations. A lower average time indicates that the system is more efficient in completing tasks quickly, which is crucial for performance in real-world applications.

\item \textbf{Survival rate:} The survival rate (\%) represents the percentage of agents that survive after performing multiple execution rounds. In out-of-distribution (OOD) scenarios, agents may face unseen dangers (e.g., Task 7), which can threaten their survival. Therefore, we use the survival rate as a key metric to evaluate robustness and adaptability in such challenging conditions. 

\item \textbf{Completion rate:} The Completion Rate measures the percentage of subtasks successfully executed by an agent relative to the total number of subtasks in a given initial target. It quantifies the agent’s ability to make progress and accomplish intermediate objectives, even if the full task is not completed.

\end{enumerate}


\subsubsection{Performance on In-domain Task}

\myparatight{DSADF is effective and efficient (Object 1 and Object 2)} In in-domain tasks, we primarily evaluate our DSADF based on two key metrics: in-domain generalization and efficiency. By assessing its performance on tasks or scenarios encountered during training processing, we can verify the system's generalization and effectiveness.
Specifically, we measure in-domain generalization by the success rate of in-domain tasks and efficiency by their average execution time.
We evaluate three distinct types of reward feedback: the RL agent with Sparse Reward, the RL agent with Curiosity Reward \cite{li2020random}, the RL agent with Self-Imitation Dense Reward \cite{oh2018self}, and a novel method that integrates RL with LLM, named LINVIT \cite{zhang2024can}. We provide a detailed introduction about this baseline in Table \ref{baseline}.
\begin{table}[!t]
\caption{Baseline Description}
\centering
\footnotesize  
\renewcommand{\arraystretch}{1.2}  
\begin{tabular}{>{\centering\arraybackslash}p{0.3\linewidth}|p{0.5\linewidth}}  
\toprule
\textbf{Method} & \textbf{Description}  \\ 
\midrule[1.2pt]
RL agent with Sparse Reward  & In a sparse reward setting, the RL agent receives a reward only when it achieves an initial target. \\ 
\midrule
\makecell[c]{RL agent with Self-Imitation\\ Dense Reward \cite{oh2018self}}  
& Self-Imitation Dense Reward guides an RL agent to receive dense rewards for imitating its past successful actions or trajectories. This approach encourages the agent to replicate previously successful behaviors, reinforcing effective strategies and accelerating learning. \\ 
\midrule
\makecell[c]{RL agent with Curiosity \\ Dense Reward \cite{li2020random}}  
& Curiosity Dense Reward provides dense rewards to encourage the RL agent to explore states of the environment it has not encountered before. This approach helps the agent discover new knowledge, accelerating the learning process and improving its adaptability to unknown environments. \\ 
\midrule
LINVIT \cite{zhang2024can}  
& LINVIT is a method that combines LLMs with RL by using the LLM as a regularizer to improve sample efficiency and reduce the number of interactions required for learning.  \\ 
\bottomrule
\end{tabular}
\label{baseline}
\end{table}

In Table \ref{results}, the performance of various methods is compared in several tasks, from Task 1 to Task 4, including mining iron, crafting a wood sword, deforestation, and attacking a cow. We compare the model's performance under two distinct pretraining methods: ELLM \cite{du2023guiding} and APT \cite{liu2021behavior}.
Our proposed method, DSADF, significantly outperforms the other methods in all tasks. The details of baseline and tasks can been seen in the Appendix.
Using the ELLM pre-trained approach, DSADF achieves a TSR of 88.3\% in Task 2 (crafting a stone sword)—exceeding other methods by at least 3.33\% and completing the task in just 1523.5 s. This approach consistently shows substantial improvements in both success rate and efficiency compared to other methods, with similar performance advantages observed across additional tasks. 

\begin{table*}[h]
\caption{Performance of various RL methods is evaluated for in-domain task using TSR (\(\uparrow\)) and Times (\(\downarrow\)) metrics, where higher TSR and lower times indicate better performance. \textbf{Bold} indicates the best performance.}
\centering
\resizebox{\linewidth}{!}{
\begin{tabular}{cc|cccccccc}
\hline
\multirow{2}{*}{Pre-trained Model}          & \multirow{2}{*}{Method}                               & \multicolumn{2}{c}{Task 1}                      & \multicolumn{2}{c}{Task 2}                      & \multicolumn{2}{c}{Task 3}                      & \multicolumn{2}{c}{Task 4}                      \\ \cline{3-10} 
                                            &                                                       & TSR (\%) \textuparrow & Times (s)\textdownarrow & TSR (\%) \textuparrow & Times (s)\textdownarrow & TSR (\%) \textuparrow & Times (s)\textdownarrow & TSR (\%) \textuparrow & Times (s)\textdownarrow \\ \hline
\multirow{5}{*}{ELLM \cite{du2023guiding}}  & RL agent with Sparse Reweard                                & 78.33                 & 3167.9                  & 71.67                 & 2979.0                  & 88.33                 & 997.6                   & 91.67                 & 845.7                   \\
                                            & RL agent with Curiosity Reweard \cite{li2020random}         & 85.00                 & 2061.4                  & 85.00                 & 2423.7                  & 93.33                 & 695.4                   & \textbf{100.00}       & 652.5                   \\
                                            & RL agent with Self-Imitation Dense Reward \cite{oh2018self} & 86.67                 & 2272.7                  & 80.00                 & 1957.6                  & 95.00                 & 875.4                   & 98.33                 & 525.6                   \\
                                            & LINVIT \cite{zhang2024can}                            & 81.67                 & 2017.4                  & 78.33                 & 2227.5                  & 91.67                 & 896.4                   & \textbf{100.00}       & 607.4                   \\
                                            & \cellcolor{greyL}DSADF                                                 & \cellcolor{greyL}\textbf{88.33}        & \cellcolor{greyL}\textbf{1523.5}         & \cellcolor{greyL}\textbf{88.33}        & \cellcolor{greyL}\textbf{1648.1}         & \cellcolor{greyL}\textbf{96.67}        & \cellcolor{greyL}\textbf{457.2}          & \cellcolor{greyL}\textbf{100.00}       & \cellcolor{greyL}\textbf{446.8}          \\ \hline
\multirow{5}{*}{APT \cite{liu2021behavior}} & RL agent with Sparse Reweard                                & 73.33                 & 3028.5                  & 70.00                 & 3375,4                  & 85.00                 & 1056.3                  & 90.00                 & 1074.5                  \\
                                            & RL agent with Curiosity Reweard \cite{li2020random}         & 81.67                 & 2374.5                  & 83.33                 & 2173.5                  & 91.67                 & 705.4                   & \textbf{100.00}       & 575.2                   \\
                                            & RL agent with Self-Imitation Dense Reward \cite{oh2018self} & 83.33                 & 2007.5                  & 76.67                 & 1972.4                  & 91.67                 & 747.2                   & 96.67                 & 517.5                   \\
                                            & LINVIT \cite{zhang2024can}                            & 80.00                 & 2157.5                  & 76.67                 & 1987.6                  & 90.00                 & 606.5                   & 96.67                 & 577.6                   \\
                                            & \cellcolor{greyL}DSADF                                                 & \cellcolor{greyL}\textbf{86.67}        & \cellcolor{greyL}\textbf{1679.2}         & \cellcolor{greyL}\textbf{88.33}        & \cellcolor{greyL}\textbf{1468.3}         & \cellcolor{greyL}\textbf{95.00}        & \cellcolor{greyL}\textbf{507.5}          & \cellcolor{greyL}\textbf{100.00}       & \cellcolor{greyL}\textbf{496.2}          \\ \hline
\end{tabular}}
\label{results}
\end{table*}

\begin{figure*}
    \centering
    \includegraphics[width=1.0\linewidth]{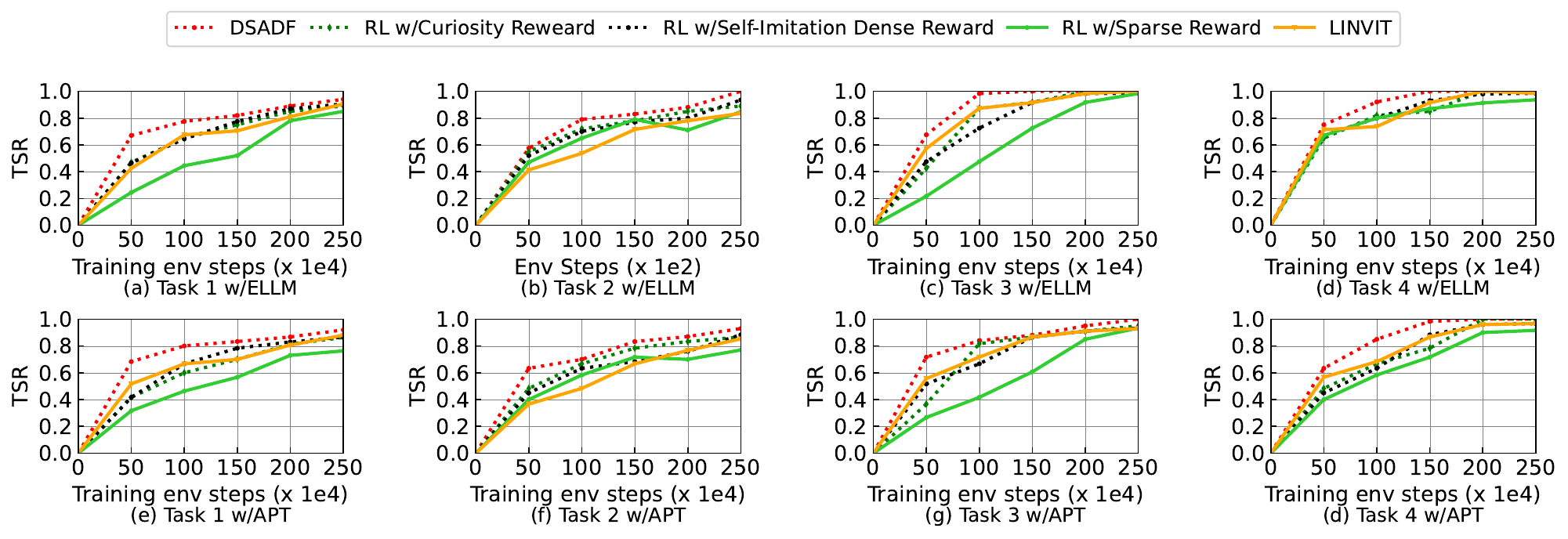} 
    \caption{The TSR trends for four downstream tasks in the Crafter environment are presented, demonstrating their progression as training steps increase. In each experiment, agents undergo pre-training using both the ELLM and APT approaches. The plots display the mean TSR, averaged over 60 independent test runs.}
    \label{fig:two_row_three_training}
     \vspace{-0.3cm}
\end{figure*}

\begin{figure*}
    \centering
    \includegraphics[width=1.0\linewidth]{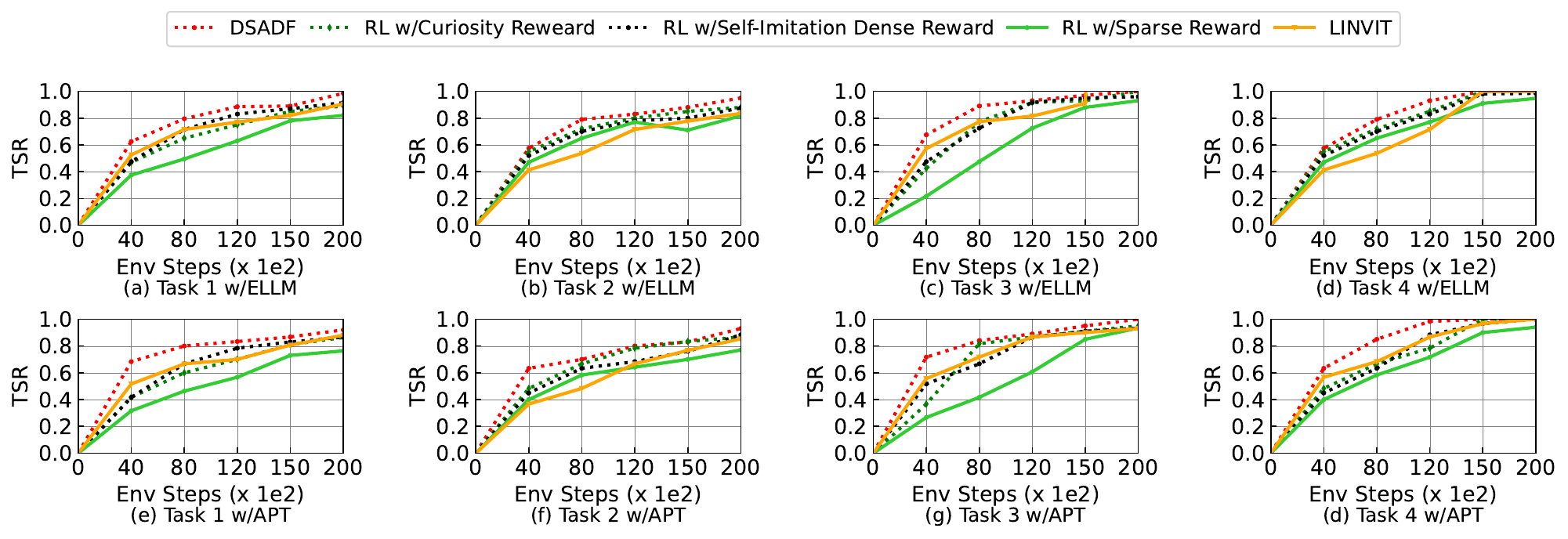} 
    \caption{TSR trends across four downstream tasks in the Crafter environment are presented, showing their progression with increasing test iterations. In each experiment, agents are pre-trained using both ELLM and APT approaches. The plots illustrate the mean TSR calculated from 60 test runs.}
    \label{fig:two_row_three_test}
     \vspace{-0.3cm}
\end{figure*}

Figure \ref{fig:two_row_three_training} shows the relationship between training steps and TSR from Task 1 to Task 4. The figure reveals that, compared to other baselines, DSADF achieves a steeper increase in TSR, indicating that it learns relevant task features more efficiently under the same number of training steps. This efficiency is attributed to our progressive reward structure, which effectively allocates rewards across different levels—from overall tasks to fine-grained action steps. As a result, our method achieves significantly faster convergence than other approaches, mitigating the optimization lag caused by the Credit Assignment Problem (CAP). This accelerates the training process and enhances overall performance.
The overall experimental results highlight that DSADF significantly outperforms other RL baselines in meeting the dual objectives of Object 1 (accurate task completion) and Object 2 (efficient task execution). Specifically, DSADF demonstrates a superior ability to complete tasks with precision while simultaneously optimizing efficiency in terms of time and resource utilization. When the evaluation is implemented under the same maximum number of steps, DSADF consistently achieves the highest TSR, indicating its strong capability to accomplish tasks accurately. Additionally, it demonstrates the shortest task completion time compared to its counterparts, underscoring its efficiency advantage. These findings suggest that DSADF strikes an effective balance between accuracy and speed, making it a robust and reliable approach for solving complex tasks within limited time constraints.
Meanwhile, Figure \ref{fig:two_row_three_test} shows the trend of TSR versus testing steps, clearly indicating that our method, DSADF, achieves a higher TSR with significantly fewer testing steps compared to other RL baselines. DSADF demonstrates a steeper trend, proving its ability to rapidly accomplish these tasks when faced with in-domain tasks, achieving the goal more efficiently (Object 2).




\begin{figure*}
    \centering
    \includegraphics[width=1.0\linewidth]{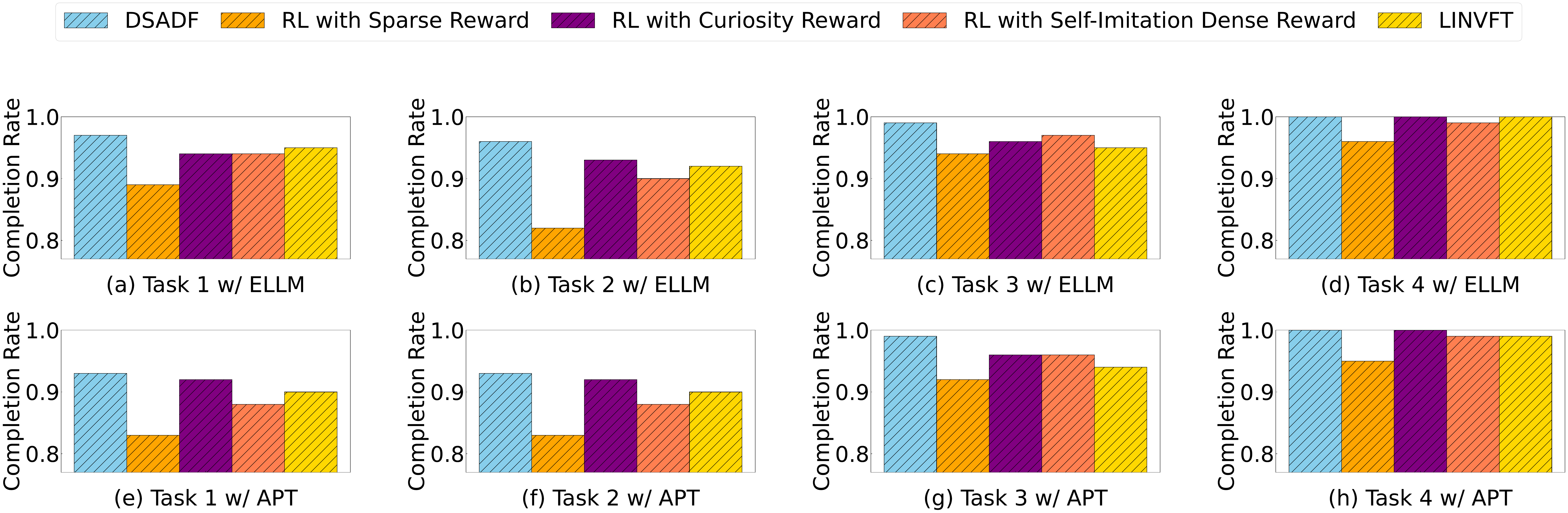} 
    \caption{Performance of various RL methods is evaluated using completion rate, where higher completion rate indicates better performance.}
    \label{fig:task1_4}
     \vspace{-0.3cm}
\end{figure*}

In addition, Figure \ref{fig:task1_4} presents the completion rates of different RL-based methods across four in-domain tasks, evaluated using two pre-trained approaches: ELLM and APT. The figure clearly demonstrates that our method achieves significantly higher completion rates compared to other baselines. This superiority holds consistently across all tasks, indicating that our approach not only excels in task success but also maintains a higher overall completion rate regardless of whether the final objective is fully achieved.
Furthermore, the results suggest that our method exhibits stronger generalization and robustness, as it outperforms competing approaches under varying task conditions. This advantage can be attributed to our framework's ability to better leverage VLM knowledge and reasoning ability which dynamically adapts to task requirements, ensuring more reliable performance.

\subsubsection{Performance on OOD Task}
\myparatight{DSADF has powerful generalization on OOD task (Object 3)} In this section, we evaluate the generalization ability of DSADF using six out-of-distribution (OOD) tasks. Since RL agents typically perform poorly on unseen tasks, we compare DSADF against three LLMs acting as agents: \textbf{LLaVA-1.5 (13B)} \cite{liu2023visual}, \textbf{MiniGPT-4 (7B)} \cite{zhu2023minigpt}, and \textbf{Qwen-2.5-VL (7B)} \cite{yang2024qwen2}. Specifically, these small VLMs act as both task planners and task executors.
In addition, we also tested a mode where a larger VLM acts as the planner and a smaller VLM as the executor. We used GPT-4o as the planner and LLaVA-1.5, Qwen-2.5-VL, and Mini-GPT4 as the executors, denoted as LLaVA-1.5$_{/\text{GPT-4o}}$, Qwen-2.5-VL$_{/\text{GPT-4o}}$, and Mini-GPT4$_{/\text{GPT-4o}}$, respectively.
The results are presented in Table \ref{unseen}, with Tasks 5 to 8 designated as long-sequence OOD tasks.  Given that GPT-4o serves as the core component of our System 2, an ideal comparison would involve larger VLMs such as GPT-4o. However, due to the substantial costs involved in using GPT-4o as an agent during testing via the API, we select three short-sequence out-of-distribution (OOD) tasks (Tasks 9 to 11, each comprising only three steps) for evaluation. The comparison results between \textbf{GPT-4o} and \textbf{Qwen-2.5 (72B)} \cite{yang2024qwen2} are presented in Table \ref{unseen_huge}.

\par Table \ref{unseen} demonstrates that our proposed DSADF method outperforms other RL agents in both generalization and efficiency on unseen tasks across all three situations. In Situation 1, DSADF achieves the highest task success rate (TSR) of 58.33\% while completing tasks in just 2767.5 seconds, significantly faster than any other agent. In general, these results highlight the robust generalization of DSADF, making it highly accurate and efficient in handling complex unseen tasks.

Table \ref{unseen_huge} compares the performance of different methods on three out-of-distribution generalization tasks (Tasks 8-10), focusing on two key metrics: TSR and completion time (Times). The experimental results demonstrate that the proposed DSADF method significantly outperforms the baselines of the large-scale vision language model (VLM) (GPT-4o and Qwen) in all tasks, exhibiting exceptional efficiency and generalizability. Specifically, on task 8, DSADF achieves a TSR of 91. 67\%, far exceeding GPT-4o (55. 00\%) and Qwen (35. 00\%), while reducing the completion time to 779.4 seconds (only one third of the baselines). In Tasks 9 and 10, DSADF further attains near-perfect success rates (98.33\%) with stable execution times under 700 seconds. In contrast, while baseline models show improved performance in simpler tasks (e.g., GPT-4o reaches 75. 00\% TSR on Task 10), they still lag significantly behind DSADF.
This superior performance likely stems from DSADF's lightweight design or domain adaptation strategies tailored for out-of-distribution scenarios, enabling high accuracy while substantially reducing computational overhead. Such advantages make DSADF a more efficient and practical solution for real-world applications.

\begin{table*}[h]
\centering
\footnotesize
\caption{Performance of out-of-distribution generalization against various foundation model baselines. TSR (\(\uparrow\)) and Times (\(\downarrow\)) are metrics, where higher TSR and lower times indicate better performance. \textbf{Bold} indicates the best performance for clarity.}
\resizebox{\textwidth}{!}{%
\begin{tabular}{c|cc|cc|ccc}
\hline
\multirow{2}{*}{Method}       & \multicolumn{2}{c|}{Task 5}                    & \multicolumn{2}{c|}{Task 6}                    & \multicolumn{3}{c}{Task 7}                                                        \\ \cline{2-8} 
                              & TSR (\%)\textuparrow & Time(s) ~\textdownarrow & TSR (\%)\textuparrow & Time(s) ~\textdownarrow & TSR (\%)\textuparrow & Time(s) ~\textdownarrow & Survival rate (\%)\textdownarrow \\ \hline
LLaVA-1.5                     & 1.67                 & 10907.3                 & 7.50                 & 9802.5                  & 35.00                & 13294.6                 & 70.00                            \\
MiniGPT-4                     & 0.00                 & 9753.6                  & 15.00                & 9563.5                  & 31.67                & 12163.5                 & 61.67                            \\
Qwen-2.5-VL                   & 7.50                 & 8495.1                  & 20.00                & 9762.4                  & 45.00                & 8751.2                  & 71.67                            \\
LLaVA-1.5$_{/\text{GPT-4o}}$   & 15.00                & 10027.6                 & 30.00                & 8358.5                  & 63.33                & 10092.7                 & 67.67                            \\
MiniGPT-4$_{/\text{GPT-4o}}$   & 7.50                 & 11274.9                 & 35.50                & 9459.7                  & 67.67                & 8229.7                  & 80.00                            \\
Qwen-2.5-VL$_{/\text{GPT-4o}}$ & 21.67                & 9397.5                  & 48.83                & 8993.5                  & 75.00                & 7452.3                  & 81.67                            \\
\cellcolor{greyL}DSADF                         & \cellcolor{greyL}\textbf{68.33}       & \cellcolor{greyL}\textbf{2767.5}         & \cellcolor{greyL}\textbf{75.00}       & \cellcolor{greyL}\textbf{1884.2}         & \cellcolor{greyL}\textbf{83.33}       & \cellcolor{greyL}\textbf{1279.9}         & \cellcolor{greyL}\textbf{93.33}                   \\ \hline
\end{tabular}
}
\label{unseen}
\end{table*}
\begin{table*}[h]
\centering
\caption{Performance of out-of-distribution generalization against larger VLM model baselines. TSR (\(\uparrow\)) and Times (\(\downarrow\)) are metrics, where higher TSR and lower times indicate better performance. \textbf{Bold} indicates the best performance for clarity.}
\resizebox{\textwidth}{!}{%
\begin{tabular}{c|cc|cc|cc}
\hline
\multirow{2}{*}{Method} & \multicolumn{2}{c|}{Task 8}                      & \multicolumn{2}{c|}{Task 9}                   & \multicolumn{2}{c}{Task 10}                   \\ \cline{2-7} 
                        & TSR (\%) \textuparrow & Times (s) \textdownarrow & TSR \textdownarrow & Times (s) \textdownarrow & TSR \textdownarrow & Times (s) \textdownarrow \\ \hline
GPT-4o                   & 55.00                 & 2745.5                  & 70.00              & 2288.5                  & 75.00              & 2459.2                  \\
Qwen                    & 35.00                 & 2792.4                  & 56.67              & 2145.7                  & 66.67              & 2023.4                  \\
\cellcolor{greyL}DSADF                    & \cellcolor{greyL}\textbf{91.67}        & \cellcolor{greyL}\textbf{779.4}          & \cellcolor{greyL}\textbf{98.33}     & \cellcolor{greyL}\textbf{687.2}          & \cellcolor{greyL}\textbf{98.33}     & \cellcolor{greyL}\textbf{674.4}          \\ \hline
\end{tabular}
}
\label{unseen_huge}
\end{table*}

\begin{figure*}
    \centering
    \includegraphics[width=1.0\linewidth]{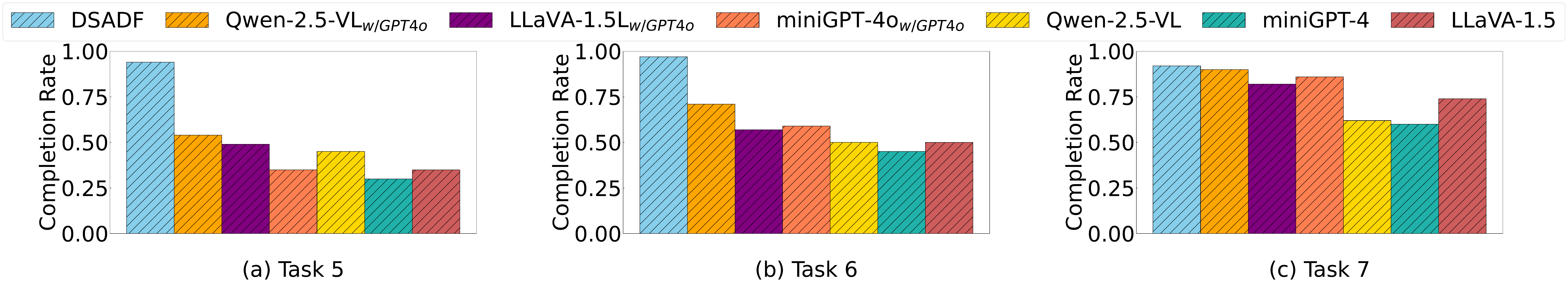} 
    \caption{Performance against various VLM-based methods is evaluated using completion rate, where higher completion rate indicates better completion.}
    \label{fig:task4_6}
     \vspace{-0.3cm}
\end{figure*}

\subsubsection{Ablation analysis}

\noindent \textbf{Impact of VLM as a planner:} Table \ref{aba_planner} employs various large-scale VLM as planners in comparative experiments. 
We employ three additional larger VLMs: Qwen-2.5 (72B) \cite{yang2024qwen2}, LLaMA-3.1 (405B) \cite{grattafiori2024llama}, and DeepSeek \cite{guo2024deepseek}.
The results demonstrate that DSADF (GPT-4o) achieves the best overall performance. Specifically, in Task 5, DSADF (GPT-4o) attains the highest TSR (68.33\%) and the shortest completion time (2767.5 seconds), significantly outperforming other planners. In Task 6, although the DeepSeek planner achieves a slightly higher TSR (77.5\%), DSADF (GPT-4o) exhibits stronger practicality with faster execution (1884.2 seconds). For the more simple Task 7, DSADF (GPT-4o) matches the highest TSR (83.33\%) alongside DeepSeek while maintaining superior time efficiency (679.9 seconds). Additionally, the Qwen-2.5 planner demonstrates the highest survival rate (94.07\%) in Task 7, albeit with longer completion times, making it more suitable for scenarios requiring higher safety. Overall, DSADF (GPT-4o) achieves the optimal balance between task success rate and execution efficiency, validating its superiority in multi-task decision-making and planning.

\begin{table*}[h]
\centering
\caption{The impact of various larger VLMs as planners on the performance of DSADF. TSR (\(\uparrow\)) and Times (\(\downarrow\)) are metrics, where higher TSR and lower times indicate better performance. \textbf{Bold} indicates the best performance for clarity. }
\resizebox{\textwidth}{!}{\begin{tabular}{c|cc|cc|ccc}
\hline
\multirow{2}{*}{Method}   & \multicolumn{2}{c|}{Task 5}                  & \multicolumn{2}{c|}{Task 6}                  & \multicolumn{3}{c}{Task 7}                                                    \\ \cline{2-8} 
                          & TSR (\%)\textuparrow & Time(s)\textdownarrow & TSR (\%)\textuparrow & Time(s)\textdownarrow & TSR (\%)\textuparrow & Time(s)\textdownarrow & Survival rate (\%)\textuparrow \\ \hline
Planner as DeepSeek       & 65.50                & 3022.7                & \textbf{77.50}       & 2458.5                & \textbf{83.33}       & 677,8                 & 90.17                          \\
Planner as Qwen-2.5       & 59.50                & 2998.4                & 68.00                & 2029.7                & 78.00                & 762.7                 & \textbf{94.07}                 \\
Planner as LLaMA-3.1      & 64.33                & 3227.5 s               & 72.33                & 2794.5                & 81.00                & 745.6                 & 92.50                          \\
\cellcolor{greyL}Planner as GPT-4o (DSADF) & \cellcolor{greyL}\textbf{68.33}       & \cellcolor{greyL}\textbf{2767.5}       & \cellcolor{greyL}75.00                & \cellcolor{greyL}\textbf{1884.2}       & \cellcolor{greyL}\textbf{83.33}       & \cellcolor{greyL}\textbf{679.9}        & \cellcolor{greyL}93.33                          \\ \hline
\end{tabular}}
\label{aba_planner}
\end{table*}

\noindent \textbf{Impact of VLM as a auxiliary performer:}
Table \ref{aba_aux} compares the performance of four small VLMs serving as auxiliary performers in the DSADF framework when handling OOD tasks (Tasks 4-6). The results demonstrate that Qwen-2.5-VL (DSADF) achieves the most robust performance across these OOD scenarios, attaining the highest or most competitive task success rates (TSR) while maintaining strong time efficiency. Specifically, it delivers the best TSR in tasks 4 (68.33\%) and 5 (75. 00\%), along with the fastest completion times for tasks 5 (1884.2 s) and 6 (679.9 s). While other models show specialized strengths, such as LLaMA-3.1's superior speed in Task 4 (2624.5 s) and miniGPT's peak TSR in Task 6 (85.00\%) - their performance proves to be less consistent across different OOD challenges. In particular, Qwen-2.5-VL combines high success rates with efficient processing times and the best survival rate (93.33\% in Task 6), suggesting that it maintains more reliable operation when faced with unexpected distribution changes. These findings highlight that auxiliary performer selection for OOD tasks requires careful consideration of both task success likelihood and computational efficiency, with Qwen-2.5-VL emerging as the most balanced solution for robust out-of-distribution performance in the DSADF framework. The varying results across tasks also underscore how different OOD scenarios may stress distinct model capabilities, warranting task-specific evaluation in practical applications.

\begin{table*}[h]
\centering
\caption{The impact of various small VLMs as auxiliary performer on the performance of DSADF. TSR (\(\uparrow\)) and Times (\(\downarrow\)) are metrics, where higher TSR and lower times indicate better performance. \textbf{Bold} indicates the best performance for clarity. }
\label{aba_aux}
    \resizebox{\textwidth}{!}{ 
    \begin{tabular}{c|cc|cc|ccc}
        \hline
        \multirow{2}{*}{Method} & \multicolumn{2}{c|}{Task 4} & \multicolumn{2}{c|}{Task 5} & \multicolumn{3}{c}{Task 6} \\ 
        \cline{2-8}
        & TSR (\%)$\uparrow$ & Time (s)$\downarrow$ & TSR (\%)$\uparrow$ & Time (s)$\downarrow$ & TSR (\%)$\uparrow$ & Time (s)$\downarrow$ & Survival rate (\%) \\
        \hline
        Aux. performer w/ LLaVA-1.5 & 63.33 & 2974.7 & 71.67 & 1973.7 & 73.33 & 722.8 & 91.67 \\
        Aux. performer w/ miniGPT & 60.00 & 2808.7 & 71.67 & 1957.8 & \textbf{85.00} & 700.4 & 93.33 \\
        Aux. performer w/ LLaMA-3.1 & 61.67 & \textbf{2624.5} & \textbf{75.00} & 2094.5 & 81.67 & 725.6 & 90.00 \\
        \cellcolor{greyL}Aux. performer w/ Qwen-2.5-VL (DSADF) & \cellcolor{greyL}\textbf{68.33} & \cellcolor{greyL}2767.5 & \cellcolor{greyL}\textbf{75.00} & \cellcolor{greyL}\textbf{1884.2} & \cellcolor{greyL}83.33 & \cellcolor{greyL}\textbf{679.9} & \cellcolor{greyL}\textbf{93.33} \\
        \hline
    \end{tabular}
    }
    \label{aba_aux}
\end{table*}

\noindent \textbf{Impact of each component:} To more clearly highlight the contribution of each component in DSADF to the final performance, we designed five distinct variations. Variation I omits the larger VLM as the planner, making it equivalent to an RL agent with sparse rewards. Variation II removes the small VLM as the auxiliary performer, relying solely on System 2 to guide the RL agent in completing the task. Variation III excludes the RL agent entirely, with only the planner guiding the auxiliary performer to accomplish the task. Variation IV eliminates both the larger VLM as the planner and the RL agent, leaving the small VLM as the sole auxiliary performer executing the task. Finally, Variation V removes the concept of self-reflection from the original DSADF framework. We will test the impact of each component separately on in-domain tasks (task 1 to task 4) and OOD tasks (task 5 to task 8) as below.

\begin{itemize}
\item { Ablation analysis on in-domain task:}  Table \ref{aba_1} demonstrate the critical contributions of each component in DSADF's hierarchical architecture for in-domain tasks. The full DSADF framework achieves optimal performance across all tasks (88.33-100\% TSR with fastest completion times), validating its integrated design. Notably, removing the large VLM planner (Variation I) causes significant TSR drops (e.g., 16.66\% decrease in Task 2), revealing its essential role in high-level task decomposition. The small VLM performer proves equally vital, as its absence (Variation II) leads to 2-3x longer execution times, highlighting its efficiency in action generation. The dramatic performance collapse in Variation III (TSR plunging to 33.75-75.5\%) confirms RL's irreplaceable role in handling complex control sequences, while Variation IV's poor results (13.33-61.67\% TSR) underscore the necessity of the full hierarchical system. Interestingly, self-reflection removal (Variation V) shows negligible impact on in-domain tasks, suggesting this mechanism primarily aids out-of-distribution adaptation. These findings collectively establish that DSADF's strength lies in the synergistic combination of the large VLM's planning capability, small VLM's rapid execution, and RL's precise control, with each component addressing distinct aspects of the problem-solving pipeline. The results particularly emphasize that while VLMs provide crucial cognitive capabilities, RL remains indispensable for physical task execution, and that the complete system outperforms any partial implementation.

\item { Ablation analysis on OOD task:} Table \ref{aba_2} compares the performance of DSADF and its five variations on out-of-domain tasks (Tasks 5–7), measuring Task Success Rate (TSR, higher is better) and completion time (lower is better). The full DSADF (highlighted in gray) consistently outperforms all variations, achieving the highest TSR (68.33–83.33\%) and fastest times (679.9–2767.5 s), demonstrating the necessity of its integrated components. Variation V (no self-reflection) performs closest to DSADF in TSR but with slightly slower times, suggesting reflection boosts efficiency without drastically altering success rates. Variation III (no RL agent) shows moderate TSR but extremely high times, revealing RL’s critical role in speed. Variations I and II (missing planner or auxiliary VLM) fail in Tasks 5–6, highlighting their importance for OOD generalization. Variation IV (only small VLM) struggles with low TSR and high times, proving hierarchical design is essential. The results underscore that DSADF’s strength lies in its synergistic combination of planning, execution, and reflection.
\end{itemize}

\begin{table*}[h]
\centering
\caption{The impact of each component on the performance of DSADF for in-domain tasks. TSR (\(\uparrow\)) and Times (\(\downarrow\)) are metrics, where higher TSR and lower times indicate better performance. \textbf{Bold} indicates the best performance for clarity.}
\label{aba_1}
    \resizebox{\textwidth}{!}{ 
    \begin{tabular}{c|cccccccc}
\hline
\multirow{2}{*}{Method} & \multicolumn{2}{c}{Task 1}            & \multicolumn{2}{c}{Task 2}           & \multicolumn{2}{c}{Task 3}           & \multicolumn{2}{c}{Task 4}          \\ \cline{2-9} 
                        & TSR (\%)$\uparrow$ & Time (s)         & TSR (\%)$\uparrow$ & Time (s)        & TSR (\%)$\uparrow$ & Time (s)        & TSR (\%)$\uparrow$ & Time (s)       \\ \hline
Variation I             & 78.33              & 3167.94          & 71.67              & 2979.60         & 88.33              & 997.65          & 91.67              & 845.73         \\
Variation II            & \textbf{88.33}     & \textbf{1523.57} & 81.67              & 2027.5          & \textbf{96.67}     & \textbf{457.29} & \textbf{100.00}    & \textbf{446.8} \\
Variation III           & 33.75              & 8943.5           & 38.33              & 7227.6          & 73.33              & 7500.4          & 75.50              & 5997.6         \\
Variation IV            & 13.33              & 13540.6          & 16.67              & 12790.4         & 28.33              & 4642.5          & 61.67              & 6397.5         \\
Variation V             & \textbf{88.33}     & \textbf{1523.57} & 80.00              & 1885.3          & \textbf{96.67}     & \textbf{457.2}  & \textbf{100.00}    & \textbf{446.8} \\
\cellcolor{greyL}DSADF                   & \cellcolor{greyL}\textbf{88.33}     & \cellcolor{greyL}\textbf{1523.57} & \cellcolor{greyL}\textbf{88.33}     & \cellcolor{greyL}\textbf{1648.3} & \cellcolor{greyL}\textbf{96.67}     & \cellcolor{greyL}\textbf{457.2}  & \cellcolor{greyL}\textbf{100.00}    & \cellcolor{greyL}\textbf{446.8} \\ \hline
\end{tabular}
    }
\end{table*}

\begin{table*}[h]
\centering
\caption{The impact of each component on the performance of DSADF for out-of-domain tasks. TSR (\(\uparrow\)) and Times (\(\downarrow\)) are metrics, where higher TSR and lower times indicate better performance. \textbf{Bold} indicates the best performance for clarity.}
\resizebox{\textwidth}{!}{\begin{tabular}{c|ccccccc}
\hline
\multirow{2}{*}{Method} & \multicolumn{2}{c}{Task 5}           & \multicolumn{2}{c}{Task 6}           & \multicolumn{3}{c}{Task 7}                                         \\ \cline{2-8} 
                        & TSR (\%)$\uparrow$ & Time (s)        & TSR (\%)$\uparrow$ & Time (s)        & TSR (\%)$\uparrow$ & Time (s)       & Survival rate (\%)$\uparrow$ \\ \hline
Variation I             & 0.00               & -               & 0.00               & -               & 0.00               & -              & -                            \\
Variation II            & 0.00               & \textbf{-}      & 0.00               & -               & 66.67              & \textbf{624.2} & 78.83                        \\
Variation III           & 21.67              & 9397.5          & 48.33              & 8993.5          & 75.00              & 7452.3         & 81.67                        \\
Variation IV            & 7.50               & 8495.1          & 20.00              & 9762.4          & 45.00              & 8751.2         & 71.67                        \\
Variation V             & 65.00              & 2884.7          & 66.67              & 1994.4          & \textbf{83.33}     & 692.4          & \textbf{95.00}               \\
\cellcolor{greyL}DSADF                   & \cellcolor{greyL}\textbf{68.33}     & \cellcolor{greyL}\textbf{2767.5} & \cellcolor{greyL}\textbf{75.00}     & \cellcolor{greyL}\textbf{1884.2} & \cellcolor{greyL}\textbf{83.33}     & \cellcolor{greyL}679.9          & \cellcolor{greyL}93.33                        \\ \hline
\end{tabular}}
\label{aba_2}
\end{table*}

\subsection{HouseKeep}
\label{house_keep}

In HouseKeep, we evaluate the agent's environmental generalization capability in handling untargeted situations. Specifically, we consider a scenario in which the agent encounters an environmental state without a predefined initial target or explicit task instructions. In such cases, the agent must leverage prior knowledge of the reward structure of the environment and use inference clues (instruction hints) to identify and perform the appropriate task. 

\textbf{Environment description:} Housekeep~\cite{kant2022housekeep} is an embodied intelligence benchmark to evaluate common sense reasoning in the home environment.
The agent would determine the correct placement of items based on common-sense object-receptacle pairings, as the environment does not provide explicit guidance. 
Therefore, the agent needs to analyze the reward conditions resulting from the changes in the environment state caused by its actions to identify the key strategy to solve the game. In the Housekeep environment, we prioritize enhancing the detection model's ability to handle unknown tasks through advanced reasoning and background knowledge integration.
In the Housekeep environment, compared to the Crafter environment, the agent can only perform low-level actions, which include moving forward, turning, looking up or down, and picking up or placing objects. Unlike Crafter, Housekeep does not provide longer action-step trajectories.


\textbf{Evaluation metric:} In this study, we use average object success rate (AOSR) as a key metric to evaluate the agent's ability to rearrange objects. The variable $N_\mathrm{CP}$ denotes the number of correctly placed objects, and the variable $N_{T}$ denotes the total number of objects. The object success (OS) measures the proportion of objects that are correctly placed by the agent in a single trial, calculated using the formula \( \text{OS} = \dfrac{N_\mathrm{CP}}{N_{T}} \times 100\% \). Having a success rate of 100\% indicates that the agent has correctly located and placed all misplaced objects at their appropriate locations. To ensure the robustness of the evaluation, we conduct \( N \) independent trials and compute the average OSR across all trials to obtain the AOSR, given by the formula \( \text{AOSR} = \frac{1}{N} \sum_{i=1}^{N} \text{OS}_i \), where \( \text{OS}_i \) denotes the OS of the \( i \)-th trial. A higher OS and AOSR indicate better agent performance in accurately placing objects, reflecting improved generalization capabilities.In this experiment, we performed 100 independent test trials for each task, with the final AOSR calculated as the mean value across all trials.

%

%
%


\textbf{Experiment setting:}  To evaluate the performance of DSADF in handling unknown target situations, we design three distinct scenarios from task 11 to task 13. Does each scenario consist of a single room containing several misplaced items and several potential receptacle options, with the detailed information provided in Appendix \ref{task_details}.
While all objects and room layouts in each task are familiar to the agent, the specific combinations of misplaced objects are presented for the first time. This novel arrangement poses a robustness challenge for the RL agent.

In such situations, defining an initial target for DSADF is challenging. To determine task completion targets, DASDF leverages the VLM's strong background knowledge and reasoning capabilities.
To address this problem, the RL agent interacts with the environment multiple times. The resulting sampled history of steps, $h_t (s_1, a_1, r_1, ..., s_t, a_t, r_t)$, containing actions, rewards, and environment states, is provided as an instruction hint to System 2, assisting DSADF in identifying the initial targets at the training stage.


%

\begin{table*}[!t]
\centering
\caption{Comparison of AOSR in Housekeep
task, and \textbf{Bold} indicates the best performance for clarity.}
\label{tab:comparison_results}
\begin{tabular}{cccc}
\midrule[1.2pt]
\multicolumn{1}{c|}{Method}                              & Task 11        & Task 12        & Task 13        \\ \midrule
\multicolumn{4}{c}{RL method}                                                                               \\ \midrule
\multicolumn{1}{c|}{RL agent with Sparse Reward}               & 71.17          & 78.95          & 75.62          \\
\multicolumn{1}{c|}{LINVIT \cite{zhang2024can} }                              & 87.97          & 85.55          & 84.50          \\ \midrule
\multicolumn{4}{c}{LLM as agent}                                                                            \\ \midrule
\multicolumn{1}{c|}{GPT-4o}                               & 74.85          & 85.25          & 70.30          \\
\multicolumn{1}{c|}{Qwen-2.5}                      & 78.65          & 70.02          & 88.58          \\ \midrule
\multicolumn{4}{c}{Ours}                                                                                    \\ \midrule
\multicolumn{1}{c|}{\cellcolor{greyL} DSADF}                               & \cellcolor{greyL}\textbf{92.49} & \cellcolor{greyL}\textbf{93.37} & \cellcolor{greyL}\textbf{95.77} \\ \midrule[1.2pt]
\end{tabular}
\label{housekeep}
\end{table*}

\textbf{Performance of environmental generalization:} Table \ref{housekeep} presents the comparison of the DSADF method with various RL baselines and foundation model agents on the Housekeep task. Specifically, we compare the performance metrics across Task 11, Task 12, and Task 13.
Among the RL methods, the RL agent with sparse rewards shows relatively poor performance, while LINVIT method exhibits strong competitiveness across all three tasks, achieving a success rate of 84.50\% on Task 13.  Due to VLM of LINVIT guide the model achieve high ASOR.
For foundation model agents, GPT-4o demonstrates considerable performance on Task 12 (85.25\%), while Qwen(72B) achieves a high success rate of 88.58\% on Task 13. Although the VLM-based method demonstrates strong performance, its effectiveness is diminished due to unfamiliarity with the environment.
DSADF leverages VLM's reasoning capabilities and background knowledge, combined with the RL agent's environmental familiarity, to complete tasks accurately and efficiently, achieving over 90\% AOSR across all three tasks.
These results indicate that the DSADF method excels not only in classification performance and generalization ability but also in accurately inferring the necessary strategies and actions in complex environments, effectively addressing variations and challenges across different task scenarios, ensuring high task success rates and stable decision-making capabilities. 
This successfully validates the effectiveness of DSADF in achieving our generalization objectives.

\section{Conclusion}
\par This study proposes a DSADF framework that integrates the RL agent and VLM. The proposed idea is inspired by Kahneman's fast and slow thinking theory. The reasoning ability of VLM is a beneficial complement to RL agents. The empirical study validates the effectiveness of the proposed method in unseen tasks. To the best of our knowledge, this work is an early work scaling cognitive theory to boost the generalization ability of the agent in the RL task. In the future, it would be meaningful to explore whether this framework could generate hallucinations caused by language models~\cite{zhou2024analyzing} or be adapted to a multi-agent environment~\cite{du2023review}.

\bibliography{main}

\newpage
\appendix

\section{Experiment details}
\subsection{Experiment implementation details}
\label{appendix:details}
\subsubsection{Task details}
\label{task_details}

\myparatight{Task 1} Craft stone sword.

\noindent Process steps: 7.

\noindent Step details: Find trees $\rightarrow$ Chop trees $\rightarrow$ place crafting table $\rightarrow$ make wood pickaxe $\rightarrow$ find stone $\rightarrow$ mine stone $\rightarrow$ make stone sword.

\noindent Task description: In this task, Our aim is to make a stone sword. We start from scratch with no initial resources or tools. The entire process is divided into seven sub-steps, with the ultimate goal of crafting a stone sword. Each step is interconnected, gradually advancing the acquisition of resources and the creation of tools.

\myparatight{Task 2} Mine iron.

\noindent Process steps: 9.

\noindent Step details: Find trees $\rightarrow$ Chop trees $\rightarrow$  place crafting table $\rightarrow$ make wood pickaxe $\rightarrow$ find stone $\rightarrow$ mine stone $\rightarrow$ make stone pickaxe $\rightarrow$ find iron $\rightarrow$ mine iron.

\noindent Task description: The task ``Mine iron" is to gain iron which consists of nine steps and is relatively complex. We begin with no resources or tools, meaning every necessary item must be gathered and crafted from scratch. Each step builds upon the last, guiding us through the process of tool creation, resource collection, and ultimately mining iron ore successfully.

\myparatight{Task 3} Attack cow.

\noindent Process steps: 6.

\noindent Step details: Find trees $\rightarrow$ Chop trees $\rightarrow$  Place crafting table $\rightarrow$ Make wood sword $\rightarrow$ Find cow $\rightarrow$ Attack cow.

\noindent Task description: The task ``Attack cow" is relatively simple, with the ultimate goal being to capture a cow. To increase the chances of success, crafting a wooden sword is necessary. The entire process is divided into six steps, starting from gathering basic resources to building the wooden sword and finally approaching the cow strategically for capture.

\myparatight{Task 4} Deforestation.

\noindent Process steps: 8.

\noindent Step details: Find tree $\rightarrow$ Chop tree $\rightarrow$ Find tree $\rightarrow$ Chop tree $\rightarrow$ Find tree $\rightarrow$ Chop tree $\rightarrow$ Find tree $\rightarrow$ Chop tree.

\noindent Task description: The goal of the ``Deforestation" task is to chop down four trees. This is one of the most basic and foundational tasks in the project. Although simple, it is divided into eight steps that cover everything from approaching the trees to collecting and managing the harvested wood. The entire process involves repeating Find Tree and Chop Tree four times.

\myparatight{Task 5} Mine diamond.

\noindent Process steps: 15.

\noindent Step details: Find trees $\rightarrow$ Chop trees $\rightarrow$  Place crafting table $\rightarrow$ Make wood pickaxe $\rightarrow$ Find stone $\rightarrow$ Mine stone $\rightarrow$ Make stone pickaxe $\rightarrow$ Find iron $\rightarrow$  Mine iron $\rightarrow$ Find coal $\rightarrow$ Mine coal $\rightarrow$ Place furnace $\rightarrow$ Make iron pickaxe $\rightarrow$ Find diamond $\rightarrow$ Mine diamond.

\noindent Task description: The goal of the ``mine diamond" task is to obtain a diamond, and it consists of 14 steps. We start from an initial state with no resources or tools, making it a relatively complex and challenging task. For a RL agent, this qualifies as an unseen task, since the agent has not encountered the sub-tasks ``Make iron pickaxe" and ``mine diamond" during training. As a result, the agent must learn to generalize and plan across multiple unfamiliar steps, requiring effective exploration, resource management, and long-term decision-making.

\myparatight{Task 6} Craft iron sword.

\noindent Process steps: 12.

\noindent Step details: Find trees $\rightarrow$ Chop trees $\rightarrow$  Place crafting table $\rightarrow$ Make wood pickaxe $\rightarrow$ Find stone $\rightarrow$ Mine stone $\rightarrow$ Make stone pickaxe $\rightarrow$ Find iron $\rightarrow$ Mine iron $\rightarrow$ Mine coal $\rightarrow$ Place furnace $\rightarrow$ Make iron sword.

\noindent Task description: The objective of the ``Craft iron sword" task is to forge an iron sword. This task comprises 14 steps and starts from an initial state with no resources or tools available. For a RL agent, this represents an unseen task, as it has not encountered the specific goal of crafting an iron sword during training. Given the number of steps and the dependencies involved in resource gathering, tool crafting, and iron smelting, this task is considered relatively complex and poses a significant challenge for generalization and planning.

\myparatight{Task 7} Attack cow with with unseen dangers.

\noindent Process steps: 6.

\noindent Step details:Find trees $\rightarrow$ Chop trees $\rightarrow$  Place crafting table $\rightarrow$ Make wood sword $\rightarrow$ Find cow $\rightarrow$ Attack cow.

\noindent Task description: This task shares the same template and initial setting as Task 3, where the goal is to capture a cow. However, during the testing phase, we introduced zombies into the environment — an element that never appeared during the training process. This unexpected change introduces new dynamics and potential threats, which can significantly disrupt the RL agent’s learned policy. As a result, the agent may exhibit degraded performance, hesitation, or complete failure in achieving the original objective, highlighting its limited generalization ability when faced with novel and unforeseen environmental factors.

\myparatight{Task 8} Mine diamond.

\noindent Process steps: 4.

\noindent Step details:  Place furnace $\rightarrow$ Make iron pickaxe $\rightarrow$ Find diamond $\rightarrow$ Mine diamond.

\noindent Task description: The objective of this task is also to obtain a diamond. Although relatively short in length, it still presents a challenge for the RL agent, as both ``make iron pickaxe" and ``mine diamond" are unseen during training. Unlike tasks that start from scratch, the initial state here is more advanced — the agent already has access to a furnace and iron. However, despite this advantage, the agent must still correctly sequence its actions to craft the required tools and successfully mine the diamond, testing its ability to generalize and adapt to novel combinations of subtasks.

\myparatight{Task 9} Craft iron sword.

\noindent Process steps: 4.

\noindent Process steps: 
Make wood pickaxe $\rightarrow$ find stone $\rightarrow$ mine stone $\rightarrow$ make stone sword.

\noindent Task description: The objective of this task is also to gain an iron sword, but unlike previous tasks, it is relatively short in length. Although the goal is the same, ``Craft iron sword" remains an unseen task during training, meaning the RL agent has never directly encountered this specific sequence before. However, in this setting, the initial state is more favorable—we already have access to a crafting table and the craft action, which simplifies the task by skipping early resource collection and tool-building steps. Despite being shorter, it still requires the agent to generalize and plan effectively to achieve the goal using the available resources.

\myparatight{Task 10} Crafting the wooden sword.

\noindent Process steps: 4.

\noindent Process steps: 
Find trees $\rightarrow$ Chop trees $\rightarrow$  place crafting table $\rightarrow$ make wood sword.

\noindent Task description: The goal of this task is also to gain the wooden sword, but during testing, Craft Wooden Sword remains an unseen task for the RL agent, as it was not included in the training phase. Despite being unseen, this task is considered easy due to its short action sequence and minimal resource requirements. We begin in an initial state with no resources or tools, but the steps needed to reach the goal are straightforward, involving basic crafting operations that rely on simple and commonly encountered sub-tasks.

\myparatight{Task 11} Item adjustment I

\noindent Items to be adjusted: Spatula,
Pressure Cooker, peppermint, Plate,
Fork, Blender, Salt Shaker

\myparatight{Task 12} Item adjustment II

\noindent Items to be adjusted: lamp, sparkling water, plant, candle holder, mustard bottle

\myparatight{Task 13} Item adjustment III

\noindent Items to be adjusted: Salt Shaker, Coffee Mug, Cutting Board, Vitamin Bottle, Hand Soap, Butter Dish, Kitchen Towel

\subsubsection{Action space details}
\label{act_sub_app}

\begin{table}[h]
\centering
\label{acts_sub}
\resizebox{\textwidth}{!}{%
\begin{tabular}{|c|l|l|}
\hline
Action name & \multicolumn{1}{c|}{Target object} & \multicolumn{1}{c|}{Achievement condition} \\ \hline
Do nothing  & None    & No specific                                                          \\ \hline
Eat         & Plant, Cow & Satiety level increases                                           \\ \hline
Sleep       & None    & Energy increase                                                      \\ \hline
Find        & Tree, Stone, Coal, Iron, Diamond & The character's position matches the resource's location (e.g., stands on the resource block) \\ \hline
Attack      & Zombie, Skeleton, Cow & The target object is defeated (e.g., the target object disappears) \\ \hline
Chop        & Tree, Grass & Plant (e.g., wood, sapling) is collected                              \\ \hline
Mine        & Stone, Coal, Iron, Diamond & Resource (e.g., Stone, Coal) is added to the inventory      \\ \hline
Drink       & Water   & The hydration level increases.                                       \\ \hline
Place       & Stone, Crafting Table, Furnace, Plant & Target item is placed in the specified location \\ \hline
Make        & Tools, Weapons (e.g., pickaxe, sword) & Crafted item is added to inventory              \\ \hline
\end{tabular}}
\caption{Action categories and descriptions }
\end{table}

\subsection{Visual display}

\subsubsection{Object visualization}

\begin{figure}[]
  \begin{minipage}[b]{0.14\linewidth}
    \centering
    \includegraphics[width=\linewidth]{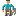}
    \caption*{Agent}
  \end{minipage}
  \hfill
  \begin{minipage}[b]{0.14\linewidth}
    \centering
    \includegraphics[width=\linewidth]{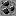}
    \caption*{Coal}
  \end{minipage}
  \hfill
  \begin{minipage}[b]{0.14\linewidth}
    \centering
    \includegraphics[width=\linewidth]{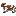}
    \caption*{Cow}
  \end{minipage}
  \hfill
  \begin{minipage}[b]{0.14\linewidth}
    \centering
    \includegraphics[width=\linewidth]{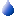}
    \caption*{Drink}
  \end{minipage}
  \hfill
  \hfill
  \begin{minipage}[b]{0.14\linewidth}
    \centering
    \includegraphics[width=\linewidth]{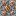}
    \caption*{Iron}
  \end{minipage}
  \hfill
  \begin{minipage}[b]{0.14\linewidth}
    \centering
    \includegraphics[width=\linewidth]{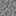}
    \caption*{Stone}
  \end{minipage}
  \hfill

  \begin{minipage}[b]{0.14\linewidth}
    \centering
    \includegraphics[width=\linewidth]{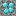}
    \caption*{Diamond}
  \end{minipage}
  \hfill
  \begin{minipage}[b]{0.14\linewidth}
    \centering
    \includegraphics[width=\linewidth]{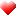}
    \caption*{Health}
  \end{minipage}
  \hfill
  \begin{minipage}[b]{0.14\linewidth}
    \centering
    \includegraphics[width=\linewidth]{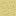}
    \caption*{Sand}
  \end{minipage}
  \hfill
  \begin{minipage}[b]{0.14\linewidth}
    \centering
    \includegraphics[width=\linewidth]{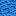}
    \caption*{Water}
  \end{minipage}
  \hfill
  \begin{minipage}[b]{0.14\linewidth}
    \centering
    \includegraphics[width=\linewidth]{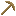}
    \caption*{Wood Pickaxe}
  \end{minipage}
  \begin{minipage}[b]{0.14\linewidth}
    \centering
    \includegraphics[width=\linewidth]{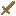}
    \caption*{Wood Sword}
  \end{minipage}

  \begin{minipage}[b]{0.14\linewidth}
    \centering
    \includegraphics[width=\linewidth]{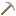}
    \caption*{Iron Pickaxe}
  \end{minipage}
  \hfill
  \begin{minipage}[b]{0.14\linewidth}
    \centering
    \includegraphics[width=\linewidth]{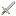}
    \caption*{Iron Sword}
  \end{minipage}
  \hfill
  \begin{minipage}[b]{0.14\linewidth}
    \centering
    \includegraphics[width=\linewidth]{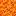}
    \caption*{Lava}
  \end{minipage}
  \hfill
  \begin{minipage}[b]{0.14\linewidth}
    \centering
    \includegraphics[width=\linewidth]{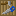}
    \caption*{Table}
  \end{minipage}
  \hfill
  \begin{minipage}[b]{0.14\linewidth}
    \centering
    \includegraphics[width=\linewidth]{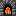}
    \caption*{Furnace}
  \end{minipage}
  \begin{minipage}[b]{0.14\linewidth}
    \centering
    \includegraphics[width=\linewidth]{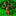}
    \caption*{Tree}
  \end{minipage}

  \begin{minipage}[b]{0.14\linewidth}
    \centering
    \includegraphics[width=\linewidth]{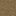}
    \caption*{Path}
  \end{minipage}
  \hfill
  \begin{minipage}[b]{0.14\linewidth}
    \centering
    \includegraphics[width=\linewidth]{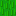}
    \caption*{Grass}
  \end{minipage}
  \hfill
  \begin{minipage}[b]{0.14\linewidth}
    \centering
    \includegraphics[width=\linewidth]{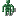}
    \caption*{Zombie}
  \end{minipage}
  \hfill
  \begin{minipage}[b]{0.14\linewidth}
    \centering
    \includegraphics[width=\linewidth]{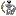}
    \caption*{Skeleton}
  \end{minipage}
  \hfill
  \begin{minipage}[b]{0.14\linewidth}
    \centering
    \includegraphics[width=\linewidth]{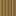}
    \caption*{Wood}
  \end{minipage}
  \begin{minipage}[b]{0.14\linewidth}
    \centering
    \includegraphics[width=\linewidth]{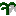}
    \caption*{Plant}
  \end{minipage}

  \caption{The object visualization of Crater. }
  \label{amv_}
  \vspace{-.15in}
\end{figure}

In Figure \ref{amv_}, we present a visualization of the main objects in Crafter.
This illustration highlights how various in-game entities are represented visually, providing insight into their design and roles. Such visualization is crucial for understanding the environment's complexity and for developing effective agent strategies.

\subsubsection{Visualization of process}

\begin{figure}[] 
  \begin{minipage}[b]{0.23\linewidth}
    \centering
    \includegraphics[width=\linewidth]{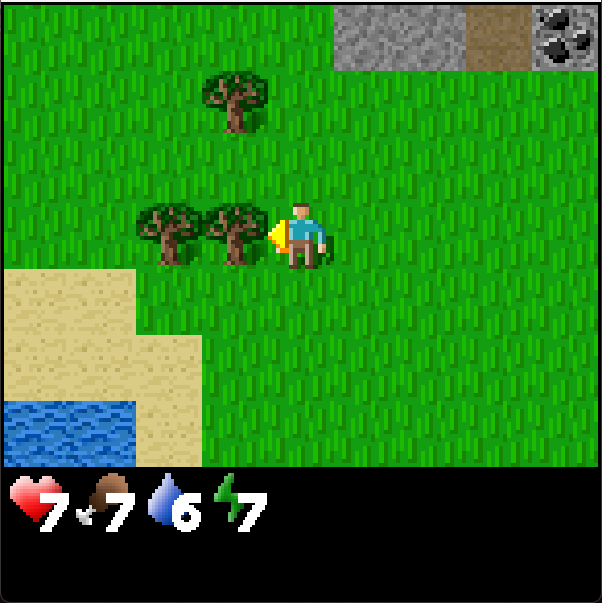}
    \caption*{Find trees}
  \end{minipage}
  \ 
  \begin{minipage}[b]{0.23\linewidth}
    \centering
    \includegraphics[width=\linewidth]{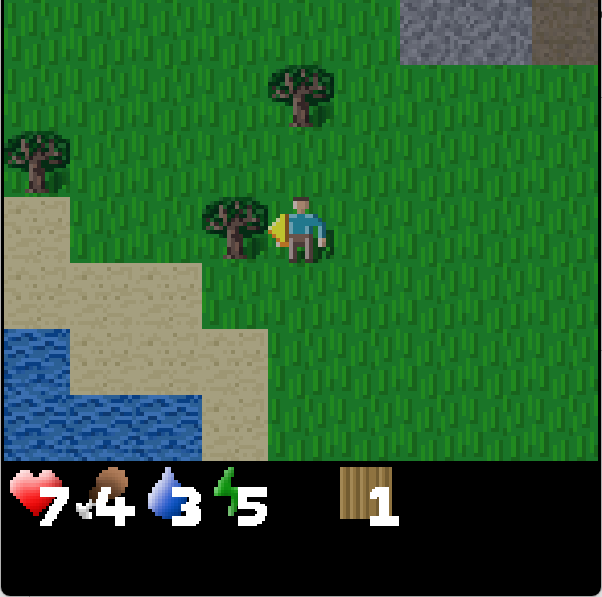}
    \caption*{Chop trees}
  \end{minipage}
   \ 
  \begin{minipage}[b]{0.23\linewidth}
    \centering
    \includegraphics[width=\linewidth]{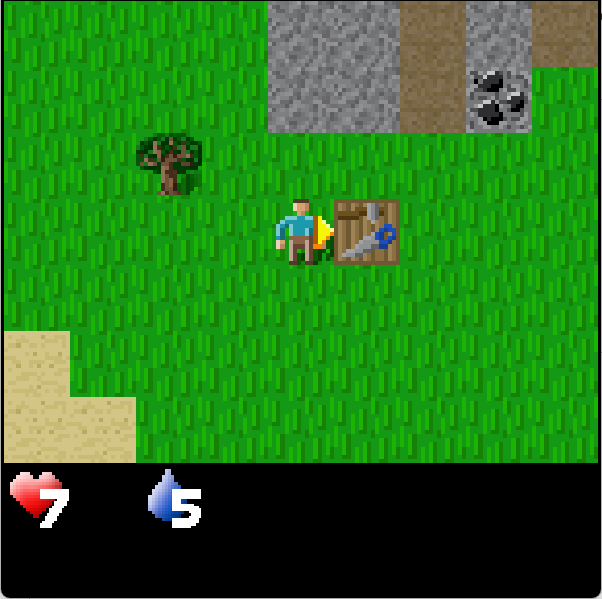}
    \caption*{Place crafting table}
  \end{minipage}
  \ 
  \begin{minipage}[b]{0.23\linewidth}
    \centering
    \includegraphics[width=\linewidth]{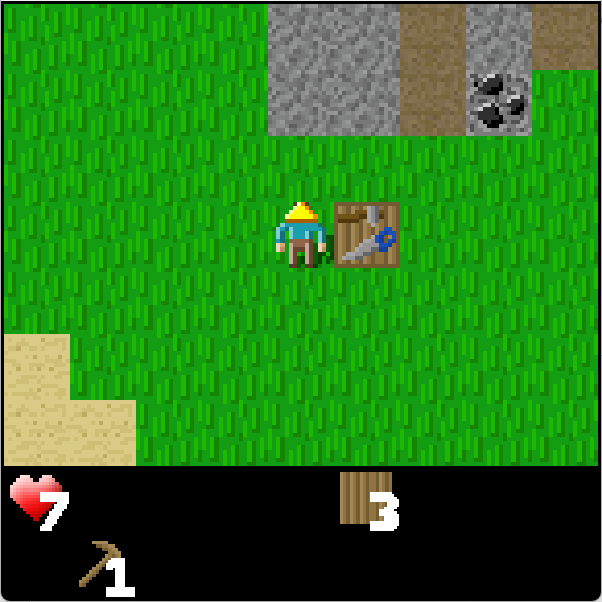}
    \caption*{Make wood pickaxe}
  \end{minipage}

  \begin{minipage}[b]{0.23\linewidth}
    \centering
    \includegraphics[width=\linewidth]{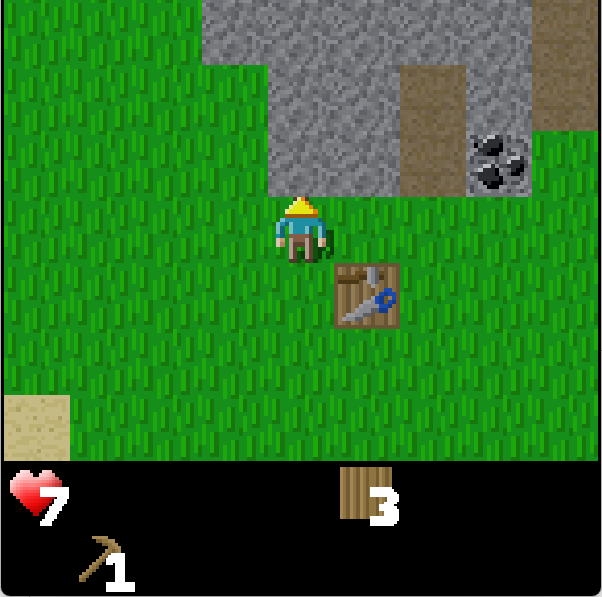}
    \caption*{Find stone}
  \end{minipage}
    \  
  \begin{minipage}[b]{0.23\linewidth}
    \centering
    \includegraphics[width=\linewidth]{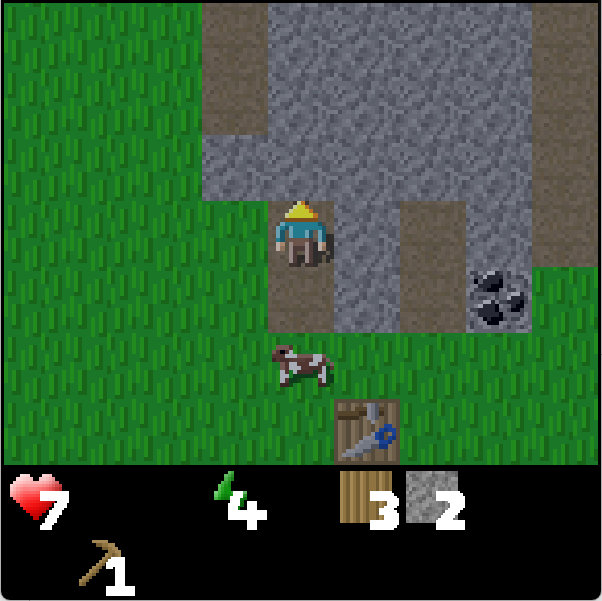}
    \caption*{Mine stone}
  \end{minipage} 
    \ 
  \begin{minipage}[b]{0.23\linewidth}
    \centering
    \includegraphics[width=\linewidth]{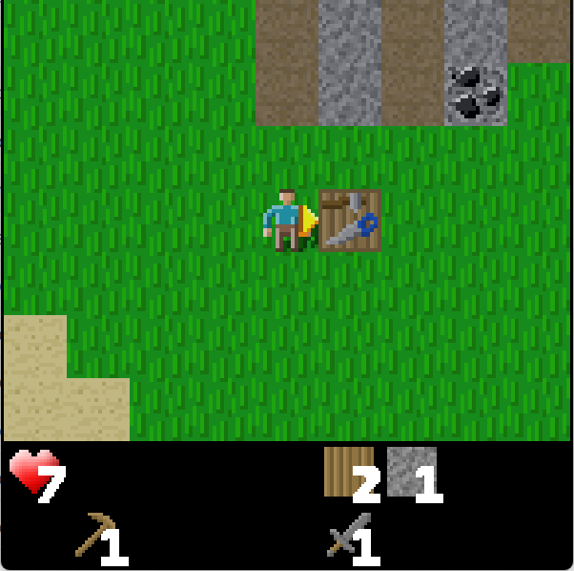}
    \caption*{Make stone sword}
  \end{minipage} 
  \begin{minipage}[b]{0.23\linewidth}
  \end{minipage}

  \caption{Visible process of crafting stone sword.}
  \label{fig:sword_crafting}
  \vspace{-.15in}
\end{figure}

\noindent \textbf{In-domain task:} Figure \ref{fig:sword_crafting} illustrates the visual process of an in-domain task (Task 1) – crafting a stone sword. All steps are performed by the RL agent.

\begin{figure}[H]
  \begin{minipage}[b]{0.23\linewidth}
    \centering
    \includegraphics[width=\linewidth]{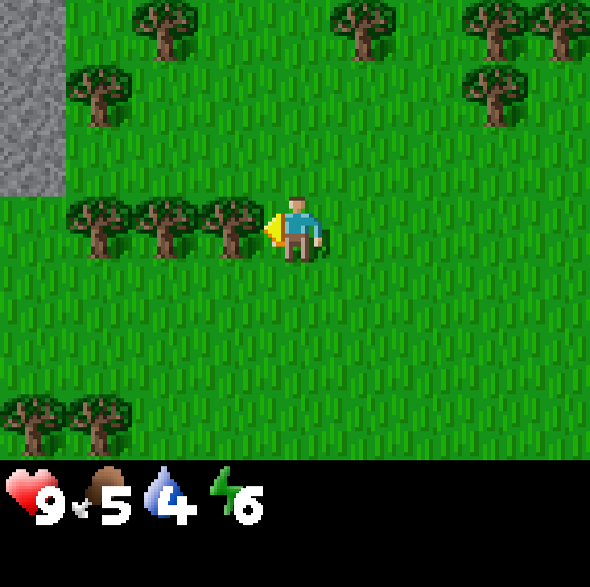}
    \caption*{Find trees}
  \end{minipage}
  \ 
  \begin{minipage}[b]{0.23\linewidth}
    \centering
    \includegraphics[width=\linewidth]{Chop_trees.png}
    \caption*{Chop trees}
  \end{minipage}
  \ 
  \begin{minipage}[b]{0.23\linewidth}
    \centering
    \includegraphics[width=\linewidth]{place_crafting_table.png}
    \caption*{Place crafting table}
  \end{minipage}
  \ 
  \begin{minipage}[b]{0.23\linewidth}
    \centering
    \includegraphics[width=\linewidth]{Make_wood_pickaxe.png}
    \caption*{Make wood pickaxe}
  \end{minipage}

  \begin{minipage}[b]{0.23\linewidth}
    \centering
    \includegraphics[width=\linewidth]{Find_stone.png}
    \caption*{Find stone}
  \end{minipage}
  \ 
  \begin{minipage}[b]{0.23\linewidth}
    \centering
    \includegraphics[width=\linewidth]{Mine_stone_2.png}
    \caption*{Mine stone}
  \end{minipage}
  \ 
  \begin{minipage}[b]{0.23\linewidth}
    \centering
    \includegraphics[width=\linewidth]{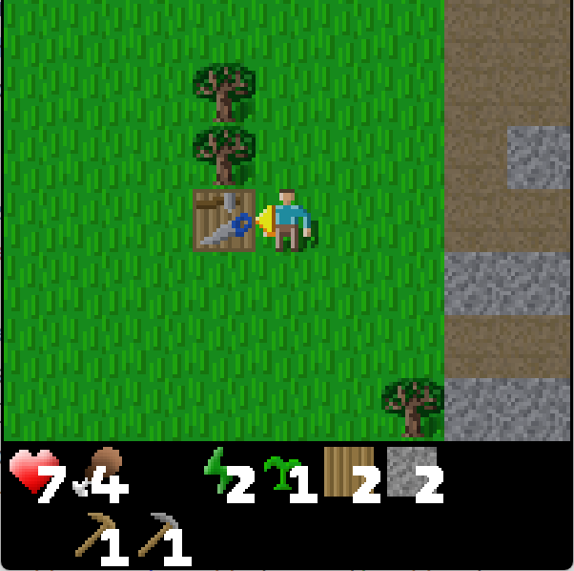}
    \caption*{Make stone pickaxe}
  \end{minipage}
  \ 
  \begin{minipage}[b]{0.23\linewidth}
    \centering
    \includegraphics[width=\linewidth]{Make_stone_pickaxe.png}
    \caption*{Find iron}
  \end{minipage}

  \begin{minipage}[b]{0.23\linewidth}
    \centering
    \includegraphics[width=\linewidth]{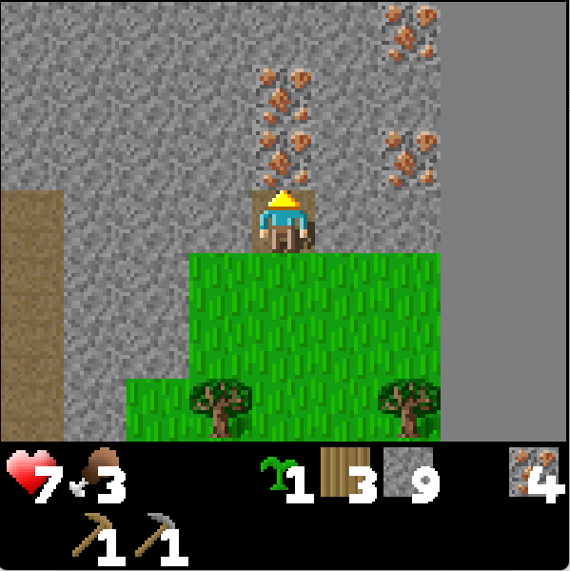}
    \caption*{Mine iron}
  \end{minipage}
  \ 
  \begin{minipage}[b]{0.23\linewidth}
    \centering
    \includegraphics[width=\linewidth]{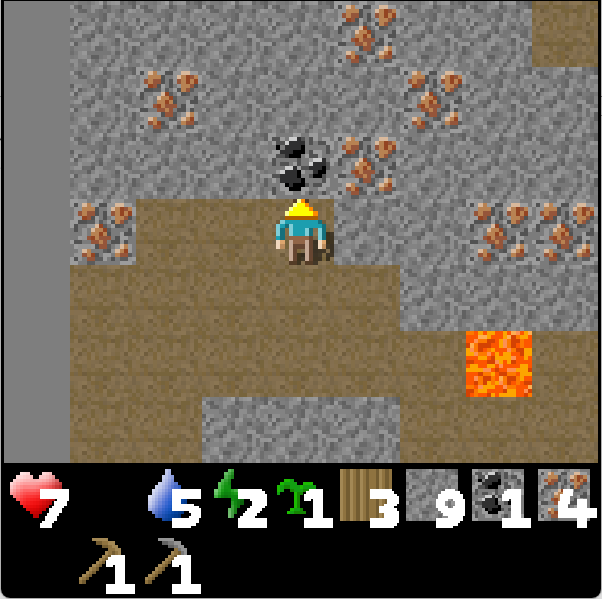}
    \caption*{Find coal}
  \end{minipage}
  \ 
  \begin{minipage}[b]{0.23\linewidth}
    \centering
    \includegraphics[width=\linewidth]{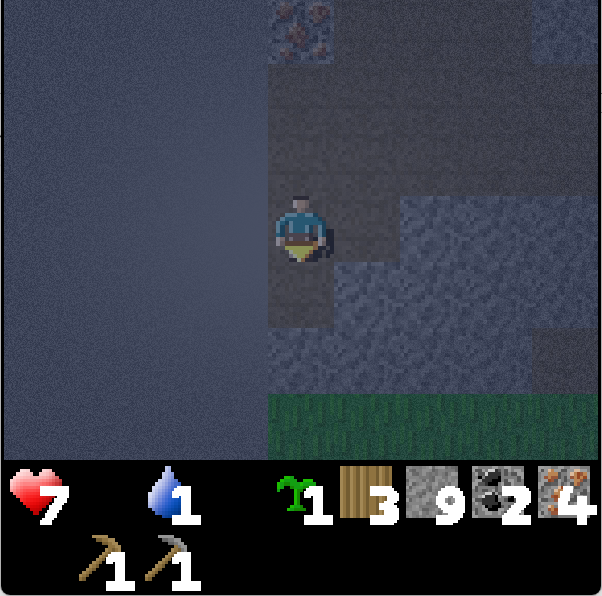}
    \caption*{Mine coal}
  \end{minipage}
  \ 
  \begin{minipage}[b]{0.23\linewidth}
    \centering
    \includegraphics[width=\linewidth]{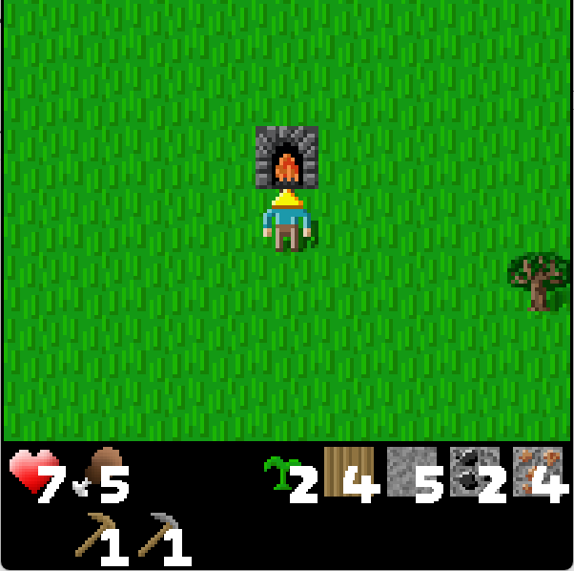}
    \caption*{Place furnace}
  \end{minipage}
  \ 
  \begin{minipage}[b]{0.23\linewidth}
  \end{minipage}

\begin{minipage}[b]{0.23\linewidth}
    \centering
    \includegraphics[width=\linewidth]{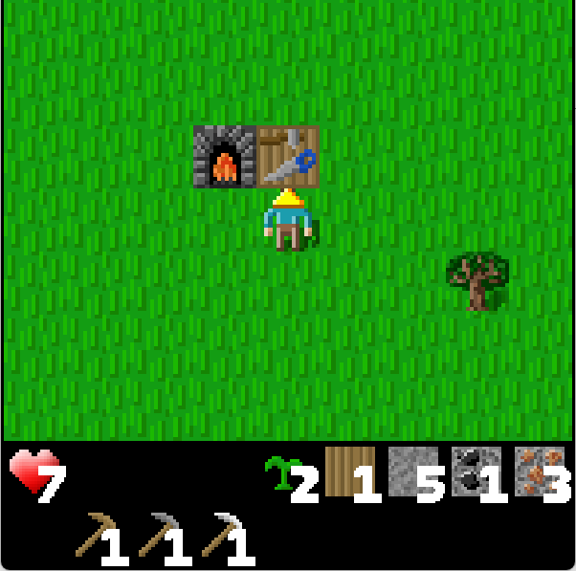}
    \caption*{Craft iron pickaxe}
  \end{minipage}
  \ 
  \begin{minipage}[b]{0.23\linewidth}
    \centering
    \includegraphics[width=\linewidth]{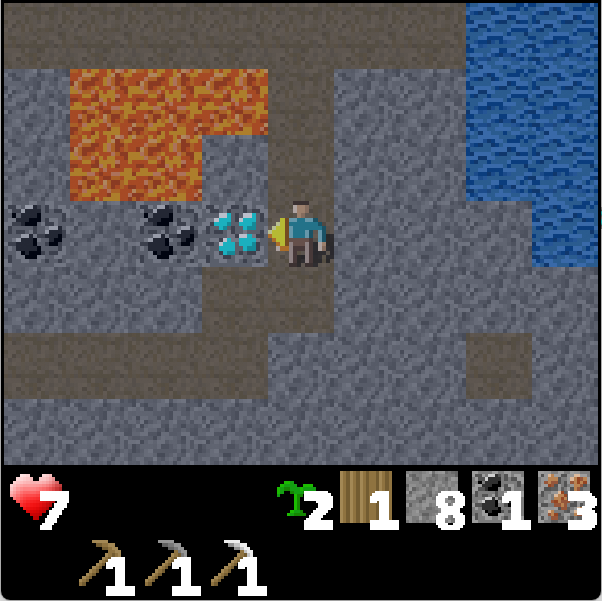}
    \caption*{Find diamond}
  \end{minipage}
  \ 
  \begin{minipage}[b]{0.23\linewidth}
    \centering
    \includegraphics[width=\linewidth]{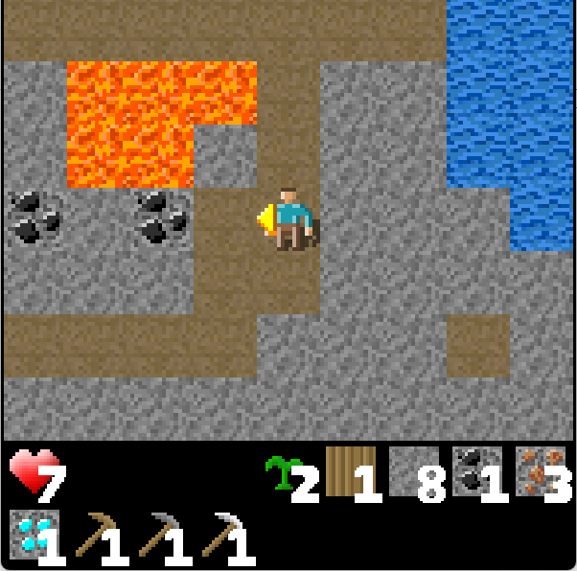}
    \caption*{Mine diamond}
  \end{minipage}
  \ 

  \caption{Visible process of Mine diamond.}
  \label{fig:dimond_crafting}
  \vspace{-.15in}
\end{figure}

\noindent \textbf{Out-of-domain task:} Figure \ref{fig:dimond_crafting} illustrates the visual process of an OOD task (Task 5) – Mine diamond. In this task, craft iron pickaxe and mine diamond performed by the VLM as the auxiliary performer, demonstrating their connectivity.

\section{Prompt of VLM}
\label{meta_promp}

\par This subsection demonstrates some prompt demos applied in our study for reference.

\subsection{VLM as planner:}

You are an intelligent task planning system for a crafter environment. Your job is to decompose high-level goals into a series of executable action steps, strictly based on a predefined action space.

You will be given:
\begin{enumerate}
    \item \textbf{Example Task} — A sample of how a goal can be decomposed.
    \item \textbf{Action Space} — A list of available actions the agent can execute.
    \item \textbf{Current Goal} — Your target task to generate a subtask plan for.
\end{enumerate}
    item \textbf{Hint instruction}- Some hint what you need.
Please follow the example format strictly. Your output should be:
\begin{itemize}
    \item Step-by-step
    \item Action-based
    \item Goal-oriented
    \item Only using actions listed in the Action Space
\end{itemize}

\subsection*{Example Task}
\textbf{Goal:} Collect wood and craft a crafting table

\textbf{Available Actions:}
\begin{itemize}
    \item \texttt{Chop Tree}
    \item \texttt{...}
    \item \texttt{...}
    \item \texttt{...}
\end{itemize}

\textbf{Example Output:}
\begin{enumerate}
    \item \texttt{Find Tree (tree\_location)} – Move to the nearest tree
    \item \texttt{...} 
    \item \texttt{...} 
    \item \texttt{...} 
\end{enumerate}

\subsection*{Action Space}
\begin{itemize}
    \item \texttt{...}
    \item \texttt{...}
    
\end{itemize}

\subsection*{Current Goal (Example)}
\textbf{Goal:} Craft a wooden sword

\textbf{Output Plan:}
\begin{enumerate}
    \item \texttt{Find tree}
    \item \texttt{Chop tree}
    \item \texttt{...}
    \item \texttt{...}
\end{enumerate}

\subsection{VLM as self-reflection}

\subsection*{Self-Reflection Prompt: Plan Evaluation}

You are a self-reflective Crafter task planner. You've just generated a step-by-step action plan to accomplish the following goal:

\textbf{Goal:} Build a simple wooden house with a door and a roof.

\textbf{Original Plan:}
\begin{enumerate}
    \item \texttt{...}
    \item \texttt{...}
\end{enumerate}

Now analyze your plan and answer the following:
\begin{itemize}
    \item Are there any missing steps, incorrect action dependencies, or unrealistic assumptions?
    \item Is the action sequence optimal in terms of efficiency and resource use?
    \item Are all required materials and tools obtained before they are used?
    \item Would an agent with the given action space be able to execute every step as written?
    \item Suggest improvements or corrections to the plan if necessary.
\end{itemize}

\textbf{Output Format:}
\begin{itemize}
    \item \textbf{Strengths:} What’s good about the plan
    \item \textbf{Weaknesses:} Potential issues or inefficiencies
    \item \textbf{Revised Plan (if needed):} A better version of the task plan
\end{itemize}

\end{document}